\documentclass[preprint,11pt,authoryear]{elsarticle}
 
\usepackage[a4paper, total={7in, 9in}]{geometry}

\usepackage{amssymb}
\usepackage{amsmath}
\usepackage{bm}
\usepackage{float}
\usepackage{url}
\usepackage[font=small,skip=-9pt]{caption}
\allowdisplaybreaks[1]

\journal{European Journal of Operational Research}

\begin{document}

\begin{frontmatter}



\title{Analysis of Fleet Modularity in an Artificial Intelligence-Based Attacker-Defender Game}


\author[label1]{Xingyu Li}		

\author[label2]{Bogdan I. Epureanu}	
\address[label1,label2]{
	Department of Mechanical Engineering\\
	University of Michigan, Ann Arbor\\
}

\begin{abstract}

Because combat environments change over time and technology upgrades are widespread for ground vehicles, a large number of vehicles and equipment become quickly obsolete. A possible solution for the U.S. Army is to develop fleets of modular military vehicles, which are built with interchangeable substantial components also known as modules. One of the typical characteristics of modules is their ease of assembly and disassembly by simple means such as plug-in/pull-out actions, which allows for real-time fleet reconfiguration to meet dynamic demands. Moreover, military demands are time-varying and highly stochastic because commanders react to adversarial actions. To capture these characteristics, we formulate an intelligent agent-based model to imitate the decision making process during fleet operation, which combines real-time optimization with artificial intelligence. Agents are capable to infer future adversarial actions based on historical data and to optimize dispatch and operation decisions accordingly. We simulate an attacker-defender game between two adversarial and intelligent fleets, one modular and the other conventional. Given the same level of resources and intelligence, we highlight the tactical advantages of fleet modularity in terms of win rate, unpredictability and damage suffered.   

\end{abstract}


\end{frontmatter}


\section{Introduction}
\label{1}



\begin{figure}
	\begin{center}
		\setlength{\unitlength}{0.012500in}%
		\includegraphics[scale=0.50]{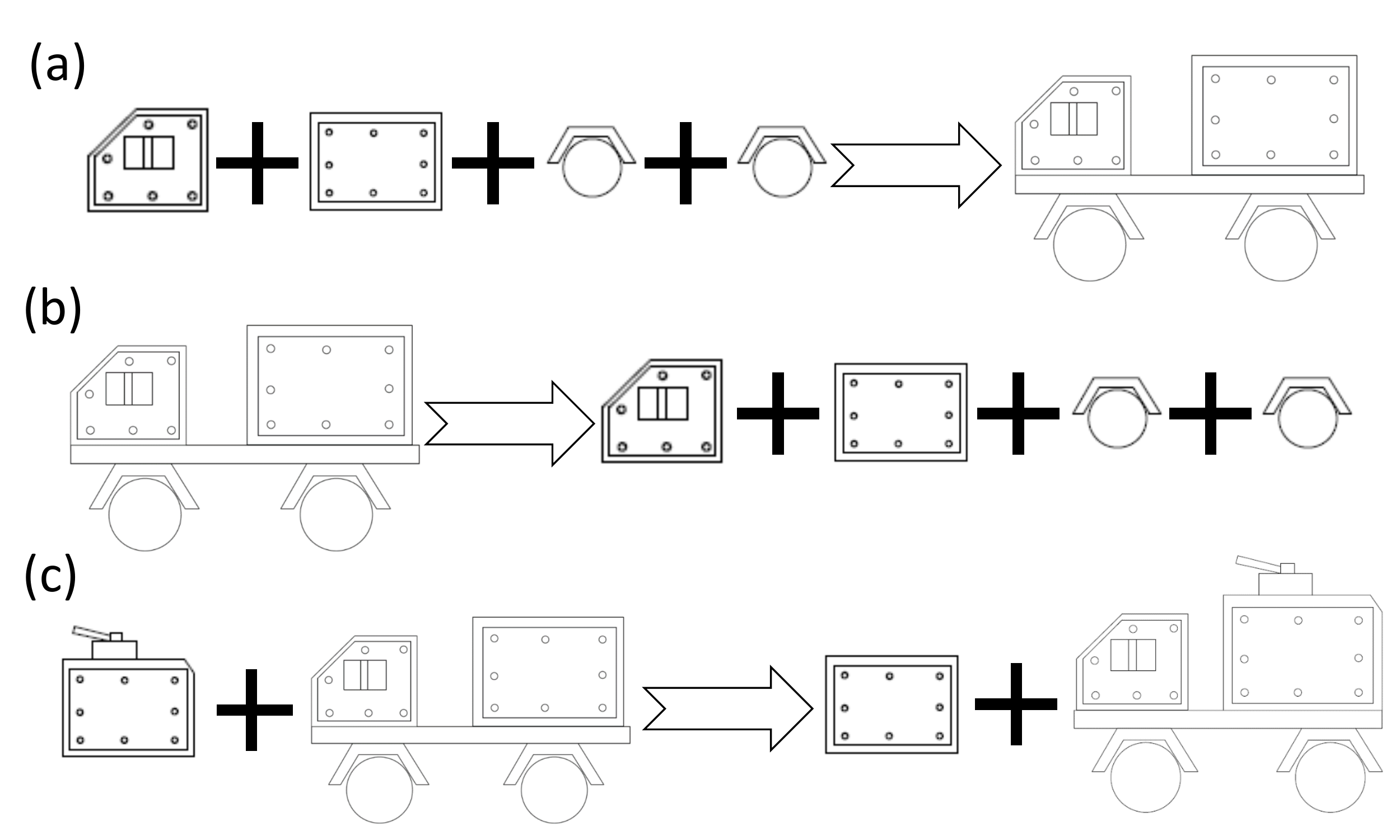}
	\end{center}
	\caption{Operation actions for modularized vehicles: a) assembly, b) disassembly, c) reconfiguration.}
	\label{fig_ADR}
\end{figure}

Military fleet operations take place in a large variety of environments and scenarios resulting in a diverse set of requirements for the fleet mix. The special functionalities of military vehicles and incessantly updated technologies make vehicles hard to reuse after military operations \citep{shinkman2014trashed}. To reduce waste, the US Army requires that fleets of vehicles can be reutilized across a large array of military mission scenarios. Modular vehicles can address this challenge \citep{dasch2016survey}. Modules are assumed to be special types of components which can be easily coupled/decoupled through simple plug-in/pull-out actions on battlefields. This property enables vehicles to be quickly assembled, disassembled and reconfigured (ADR) (as shown in Fig.~\ref{fig_ADR}) on battlefield to react to demands.

Because of the close connection between the strategy and the fleet performance, researchers have investigated the potential of modularity to boost performance during fleet operations. Bayrak et al. proposed a mathematical model to simulate modular fleet operations in a logistic mission scenario \citep{bayrak2016system}. They noticed a significant operational cost reduction after fleet modularization.  Li and Epureanu proposed an intelligent agent-based model in managing modular fleet operations \citep{li2017intelligent, li2018IABS}. Agents were classified into three categories: camp, distributor, and supply. Different types of agents collaboratively and real-timely yielded operational decisions to react to stochastic battlefield demands. Later, Li and Epureanu also modeled the fleet operation as a dynamic system and implemented model predictive control to manage the system dynamics \citep{li2017robustness}. Their results show that modular fleets exhibit a better robustness than conventional fleets as they react to disturbances from the battlefields.  

However, most previous research focused on a single decision maker who operates against a stochastic demand without intelligence. However, unpredictability from the adversarial reaction is essential and leads to an inability to forecast the outcome of actions or weakly perceived causal links between events on the battlefields \citep{Lynch2014On}. Furthermore, smart systems and artificial intelligence are playing an ever-increasing role. Autonomous vehicles, especially unmanned aerial vehicles, have been widely used to assist military operations \citep{landa1991war,jose2013technology,evers2014online}. It is no surprise that artificial intelligence plays a significant role in management of large-scale fleets of autonomous vehicles. Given an equivalent autonomous decision making system, our goal is to explore the synergy between modularity and autonomy by performing an attacker-defender game between a conventional fleet and a modular fleet.

The use of games in modeling the relationship between an attacker and a defender has a long history starting with the work of Dresher \citep{dresher1961games}. The variety of applications and research relate to issues in military operation research and defense studies are rich \citep{kardes2005survey,hausken2009minmax, zhuang2010modeling,paulson2016game}. There are also several studies of attacker-defender games that consider resource-dependent strategies. For example,  Powell used game theoretical approach to find a defender's resource allocation strategy for protecting sets from being destroyed by a strategic adversary, i.e., terrorist group \citep{powell2007defending}. The defender is uncertain about the targets that the adversary is likely to strike. Given the distribution of adversary's behaviors, Powell derived a pure strategy for the defender that leads to Bayesian Nash equilibria. Hausken and Zhuang considered a multi-period game where a defender can allocate resources to both defend its own resources and attack the resources of the adversary. Similarly, attacker can also determine the use of their resources for attacking the defender or protecting itself \citep{hausken2011defending}. They adopted a strategy pair which is proven to be a subgame perfect Nash equilibrium for a two-stage game, and they illustrated how the strategy depends on changes in the adversarial resources. 

Methods based on game theory are popular in the existing literature related to resource-dependent attacker-defender games. However, these applications mainly focus on single-period games or repeated games where most information from previous periods is ignored. Furthermore, strong assumptions on previous approaches, i.e., single resource type, sequential moves, perfect information of the adversary, also make previous research inapplicable to realistic military missions, where the demands and the environment are stochastic and unpredictable \citep{shinkman2014trashed, Lynch2014On, xu2016modeling}. Furthermore, a recent study also shows that the performance of a modular fleet is strongly influenced by the optimality of operation decisions \citep{li2018IABS, li2017intelligent}, which makes the analytical solution intractable. A new method is required to investigate the tactical advantages of fleet modularity.

In this study, we go beyond the existing literature and formulate an attacker-defender game using intelligent agent-based modeling techniques \citep{adhitya2007model, yu2009intelligent, onggo2016test, li2017intelligent}. We combine optimization techniques and artificial intelligence to enable each fleet(player) to forecast adversary's behaviors based on experience and to optimize decisions accordingly. By selecting one player as a modular fleet and another as a conventional fleet, we explore the benefits of fleet modularity, including adaptability and unpredictability, when competing in a hostile mission scenario.

\section{Game Formation}
\label{2}

Miltary fleets can face supply shortages due to exogenous supply chain disruptions during armed conflicts \citep{xu2016modeling}. Such events can be explored as a game involving a competition between two military fleets, a red fleet and a blue fleet. The goal of the fleet operation is to satisfy the supply demands which randomly appear at battlefields. Each demand is characterized by a due time, required materials, personnel, and target fleet to accomplish the demand. To satisfy the demand, dispatched convoys have to be formed with vehicles selected from each fleet so that they have enough capacity for delivering the supplies and enough firepower to guarantee the safety of transportation. For convenience, all demands are automatically converted into the attribute requirements for the convoy, i.e., firepower - the ability to deliver effective fire on a target, material capacity - the ability to relocate materials, personnel capacity - the ability to carry personnel, etc. We denote the demands received at time $t$ as $\bm{r}(t)$. According to the due time of demand, attributes required to be satisfied at a future time $t+\tau$ can be obtained as $\bm{d}^x(t+\tau) = [d^x_1(t+\tau),d^x_2(t+\tau),...,d^x_{N_a}(t+\tau)]^T$, where $d^x_e(t+\tau)$ represents the attributes of type $e$ to be satisfied before time $t+\tau$. $N_a$ is the number of attribute types of interest in this game . Matrix $\bm{D}^x(t)$ can also be created and updated to record the demands to be satisfied in the planning horizon $T_p$, i.e., $\bm{D}^x(t) = [\bm{d}^x(t+1),\bm{d}^x(t+2),...,\bm{d}^x(t+T_p)]$. Correspondingly, attributes carried by the dispatched convoy at time $t$ are $\bm{v}^x(t) = [v^x_1(t),v^x_2(t),...,v^x_{N_a}(t)]^T$.

The fleet specified as the target fleet by demand becomes defender, with the goal of delivering a qualified convoy to battlefield on time. The other fleet becomes the attacker automatically. The attacker dispatches convoy with the aims to disrupt the fulfillment of the demands by the defender by dispatching an attacker convoy. Thus, each demand initializes a supply task for one fleet and an attack task for another fleet. Based on the demand, the role of the player dynamically changes. The common rules of the game are assumed to be known by the players: delivery of a convoy with capacity of satisfying the demand makes the defender win and the attacker lose. Because the damage from the attacker can reduce the attributes of the defender's convoy, the defender may lose the game even if a convoy with higher attributes than requirements is dispatched by the defender.

\begin{figure}
	\begin{center}
		\setlength{\unitlength}{0.012500in}%
		\includegraphics[scale=0.35]{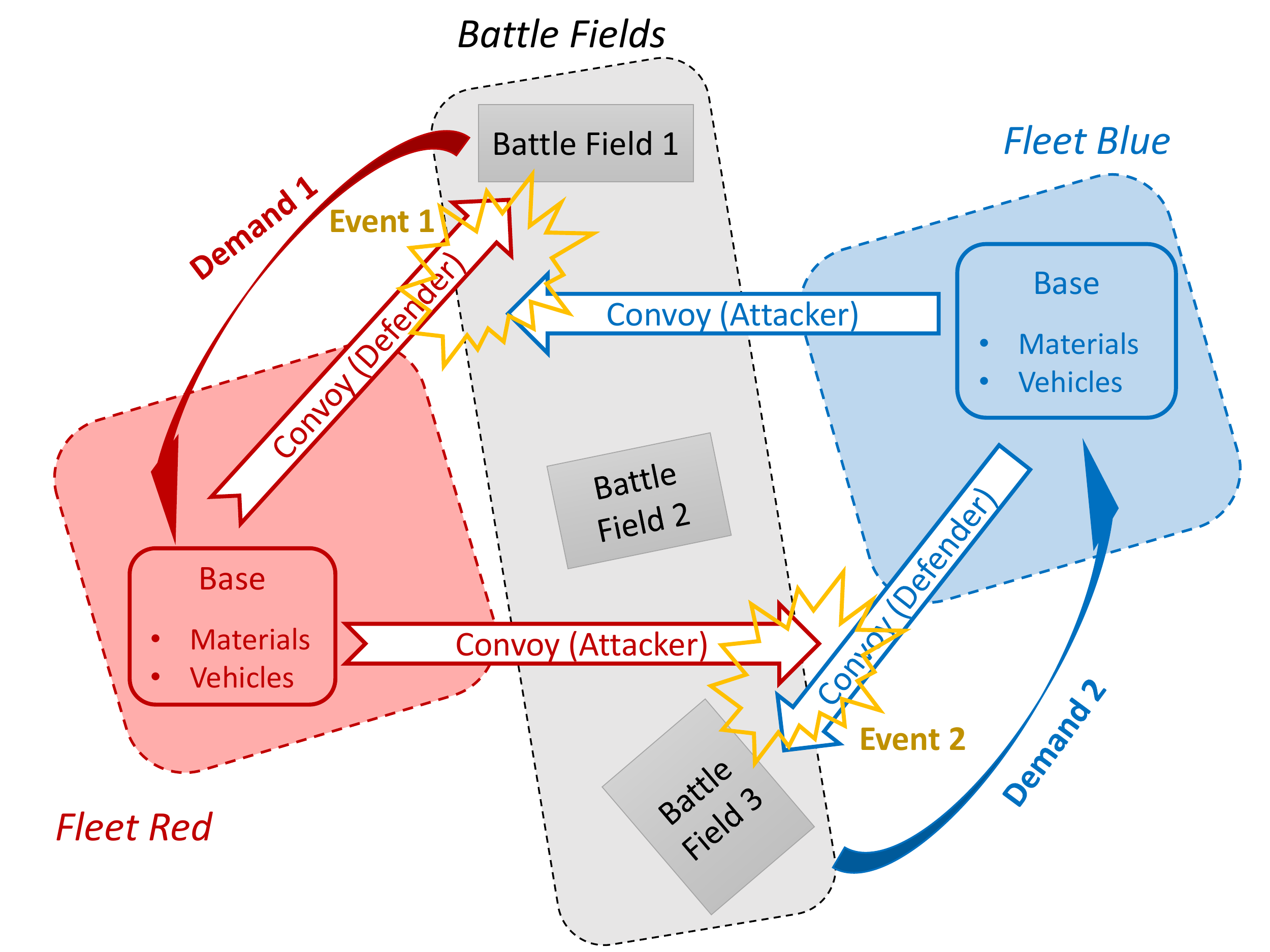}
	\end{center}
	\caption{Competition between two fleets at battlefield}
	\label{fig_bf}
\end{figure}

We denote the conflict between attacker and defender convoys as an event, and assume that the dispatch decisions made by the attacker and the defender are simultaneous. Thus, the game is a simultaneous-move game. To simplify the problem, we assume that both fleets can simultaneously sense the demands regardless of their target. Thus, given the same probability to be selected as the target fleet, each fleet will play equivalent times as attacker and defender to guarantee the fairness of the game. Fig.~\ref{fig_bf} illustrates the convoy competition in a multiple battlefield scenario. 

Following previous research \citep{azaiez2007optimal, wang2011target}, both attacker and defender are modeled as rational and strategic. Based on the simulation results, we summarized a fixed number of predefined dispatch strategies for the attacker and the defender respectively, as shown in Tab.~\ref{tab_strategies}. For an attacker, the firepower is considered as the only attribute needed to win. Thus, we classify the attack strategy by comparing the firepower to be assigned to attacker convoy to the required firepower for adversary, which is measured by $k_a$. The defender wins when it delivers a convoy which satisfies the demands with consideration of attribute losses due to adversarial disruptions. The strategy is a mixture of decisions in selecting safety coefficients of firepower $k_a$ and capacity(personnel and material) $k_c$. We cluster all the dispatch orders that are less than the requirements in strategy 1, where the defender gives up the game to save resources once a strong adversarial convoy is predicted.

\begin{table}
	\caption{Dispatch strategies for the attacker and the defender fleets}
	\label{tab_strategies}
	\begin{center}
		\begin{tabular}{c l l l l l l l l l l l}
			& & \\ 
			\hline
			\hline
			Attack Strategy  & 1 & 2 & 3 & 4 & 5\\ 
			Range of $k_a$  & $[0, 0.5)$ & $[0.5, 1)$ & $[1, 1.5)$ & $[1.5, 2)$ & $[2, 2.5)$ \\
			\hline
			Attack Strategy & 6 & 7 & 8 & 9 & 10\\
			Range of $k_a$  & $[2.5, 3)$ & $[3, 3.5)$ & $[3.5, 4)$ & $[4, 4.5)$ & $[4.5, \infty)$\\
			\hline
			\hline
			Defense Strategy  & 1 & 2 & 3 & 4 & 5\\ 
			Range of $k_a$  & $[0, 1)$ & $[1, 1.5)$ & $[1, 1.5)$ & $[1, 1.5)$ & $[1.5, 2)$ \\
			Range of $k_c$  & $[0, 1)$ & $[1, 1.5)$ & $[1.5, 2)$ & $[2, \infty)$ & $[1, 1.5)$ \\
			\hline
			Defense Strategy & 6 & 7 & 8 & 9 & 10\\
			
			Range of $k_a$   & $[1.5, 2)$ & $[1.5, 2)$ & $[2, \infty)$ & $[2, \infty)$ & $[2, \infty)$\\
			Range of $k_c$   & $[1.5, 2)$ & $[2, \infty)$ & $[1, 1.5)$ & $[1.5, 2)$ & $[2, \infty)$\\
			\hline
		\end{tabular}
	\end{center}
\end{table}

In this study, the amount of damage is based on the comparison of the firepower carried by different convoys. The probability of damage of the component of type $i$ of the red fleet $p_{di}^r$ and the blue fleet $p_{di}^b$ shown by
\begin{eqnarray}
p_{di}^r = \tanh(k_{di}\frac{v^b_{f}}{v^r_{f}})\\
p_{di}^b = \tanh(k_{di}\frac{v^r_{f}}{v^b_{f}}),
\end{eqnarray}
where $v^x_f$ is the amount of firepower carried by convoy $x$. $k_{di}$ represents the damage factor for a component in type $i$. Each component in the convoy is damaged stochastically based on the calculated probability. 

To create a fair game, we constrain the amount of supplies, and assume all damaged resources are recoverable. Thus, the amount of resources for both fleets are constant, but the conditions of the resources are dynamic. We penalize damage by requiring a long time to recover. The recovery strategy for damaged vehicles is to replace all the damaged components by healthy ones. Once vehicle modularity is considered, disassembly becomes another option in dealing with a damaged vehicle. Several assumptions are also used to simplify this problem while maintaining a reasonable level of fidelity:

\begin{enumerate}
	\item Each fleet can accurately observe and record the damage that occurs in its convoys.
	\item Each fleet can accurately observe the composition of the adversarial convoys in every event, i.e., number and types of vehicles.
	\item Convoys return to base immediately after completing a mission task. 
	\item Mission success is reported to both fleets once it is completed.
	\item No other type of vehicle damage is considered besides damage caused by the attacker.
	\item All vehicle components are recoverable.
	\item Damage occurs independently in vehicles based on the probability of damage.
	\item The inventory status is updated every hour and accessible to all the agents. 	
\end{enumerate}

\section{Problem Statement}
\label{problemStatement}
To explore the advanced tactics enabled by fleet modularity, we first consider the way operation decisions are made. With simplification, the main procedure involved in decision making \citep{chen2013human} are: 1) perceive battlefield information, 2) analyze adversary's behavior based on received information, 3) optimally schedule the operational actions to accomplish the demands or dispute adversarial actions. Through these procedures, decision makers adaptively adjust their dispatch strategy and operation plan based on the learned adversary's behaviors, i.e., what kind of strategy adversary might adopt in a specific situation. Combined with fleet modularity, a decision making process with guaranteed decision optimality and efficiency is challenge addressed herein.

Another existing challenge is the management of an inventory with high diversity. We denote a vehicle with one or more than one damaged components as damaged vehicles, which generates numerous types of damage. For example, a single vehicle with 5 different components has $2^5$ types of vehicle damage, which leads to many distinct recovery strategies. Furthermore, given the limited working capacity that can be used for ADR actions and recovery, it is still challenging to effectively and efficiently schedule operations and select the recovery strategy in reaction to stochastically arrived demands while maintaining reasonable healthy stock levels.


\section{Agent-Based Model}

{This section presents an agent-based model to automatically yield adaptive tactics and real-timely plan for operational actions accordingly. The model takes into consideration the computational load and the battlefield decision making process. To simplify the notation, we describe the approach from the standpoint of the blue fleet while the red fleet is the adversary. Three types of agents are created to perform different functionalities. The decision-making process is then achieved by the cooporation of these three types of agents. The interconnections are shown in Fig.~\ref{fig_ABM}.
	
\begin{enumerate}
	\item \textbf{Inference Agent}: analyze adversarial historical behaviors, forecast adversarial future actions. 
	\item \textbf{Dispatch Agent}: optimize dispatch order based on the output of the inference agent.
	\item \textbf{Base Agent}: optimally plan operation actions to satisfy the dispatch order.
\end{enumerate}

\begin{figure*}
	\begin{center}
		\setlength{\unitlength}{0.012500in}%
		\includegraphics[scale=0.40]{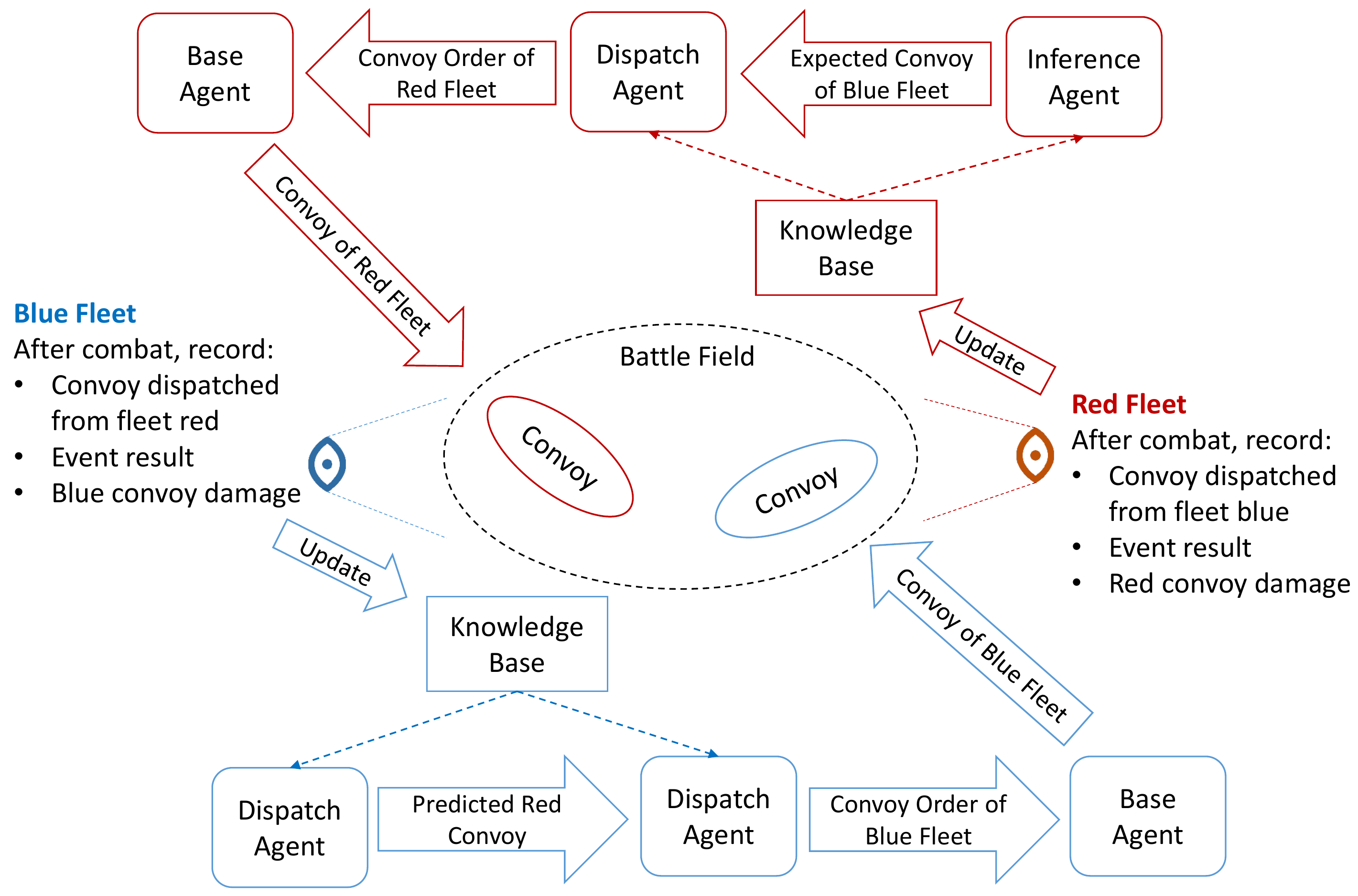}
	\end{center}
	\caption{Agent-based model in attacker-defender game}
	\label{fig_ABM}
\end{figure*}
	
} 

\subsection{Inference Agent}
The attacker-defender game is a simultaneous-move game. Hence, it is critical to forecast the adversarial actions that used to be accounted for. As combat resources and working capacity are limited, it is possible to possible to find cues from the adversarial historical dispatch actions by inference. For example, if the adversary has dispatched a convoy with a significant amount of vehicles in the near past, it is possible to conclude that the adversary is not capable of grouping up a strong convoy again in the short future. In addition, existing damage in adversary's resources can be analyzed by comparing the firepower of dispatched convoys from historical events. The amount of damage is also useful for the decision maker to infer adversary's available resources. 

The information that can be used for inference is limited, including demand records $\bm{D}(t)$, previous dispatched vehicles, $\bm{v}^b (\tau)$, and the adversary's previous dispatches $\bm{v}^r(\tau)$.  Dispatch decisions depend on an optimization algorithm, on inference of adversarial actions, and on the performances of the decision makers. Together, these lead to a strong nonlinearity in the decision-making process. In addition, the strategy employed has to be adjusted after learning from the adversarial actions. The decision-making process requires a prediction model which is able to update in real-time and capture the causality and correlation from adversary's historical behaviors. We adopt techniques from artificial intelligence to solve this problem.

\begin{figure}[h!]
	\begin{center}
		\setlength{\unitlength}{0.012500in}%
		\includegraphics[scale=0.50]{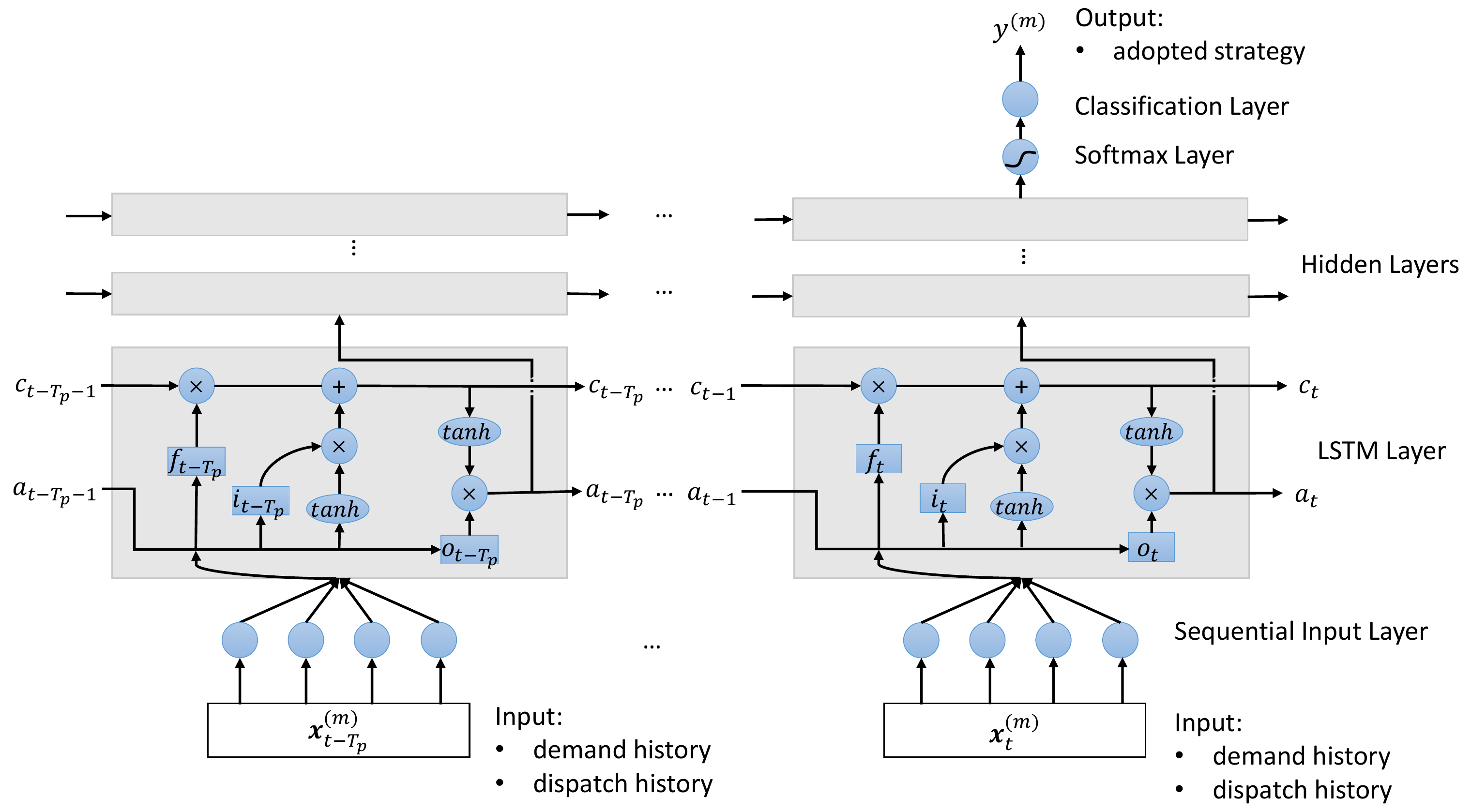}
	\end{center}
	\caption{LSTM architecture designed for inferring the adversary's strategy}
	\label{fig_training}
\end{figure}

Recurrent neural networks(RNNs) are well known machine learning models that could capture dependencies in time-series data \citep{mikolov2010recurrent}. Compared to conventional neural networks, RNNs can memorize a certain period of historical data and analyze its influence on the future. Long short-term memory model networks is one of the popular RNNs, which is capable of learning long-term dependencies without gradient vanishing problem in RNN. In this study, we implement a variant model of RNN, namely a long short-term memory (LSTM) as our predictive model to capture the correlations in adversarial sequential decisions. The LSTM model is widely used in forecasting based on sequential data, including, stock market prices \citep{chen2015lstm, di2016artificial, fischer2017deep}, traffic \citep{ma2015long, zhao2017lstm}. In this study, we model the inference of the adversary's strategy as a classification problem, where each class corresponds to a strategy. The inputs $\bm{x}^{(m)}$ of training data are the records of each event, including own dispatched convoys, adversary's dispatched convoy and the received demands, which are time-series data recorded during a time horizon $T_b$. The output $y^{(m)}$ of each training sample is the actual dispatch strategy adopted by the adversary. The architecture of LSTM used in this study is shown in Fig.~\ref{fig_training}.

The status of the LSTM model at time $t$ are described by the input gate $\bm{i}_{t}$,  forget gate $\bm{f}_{t}$, output gate $\bm{o}_{t}$, cell state $\bm{c}_{t}$ and activation at $i ^{th}$ hidden layer $\bm{a}^i_{t}$. The forward propagation can be described by
\begin{eqnarray}
& \bm{i}_{t} = \sigma (\bm{W}_i [\bm{a}_{t-1}, \; \bm{x}^{(m)}_{t}] + \bm{b}_i),\\
& \bm{f}_{t} = \sigma (\bm{W}_f [\bm{a}_{t-1}, \; \bm{x}^{(m)}_{t}] + \bm{b}_f),\\
& \bm{o}_{t} = \sigma (\bm{W}_o [\bm{a}_{t-1}, \; \bm{x}^{(m)}_{t}] + \bm{b}_o),\\
& \bm{c}_{t} = \bm{i}_{t} * \tanh (\bm{W}_c [\bm{a}_{t-1}, \; \bm{x}^{(m)}_{t}] + \bm{b}_c) + \bm{f}_{t} * \tanh (\bm{W}_c [\bm{a}_{t-2}, \; \bm{x}^{(m)}_{t-1}] + \bm{b}_c),\\
& \bm{a}^0_{t} = \bm{o}_{t} * \tanh({\bm{c}}_{t}),\\
& \bm{a}^1_{t} = \bm{W}_{1}\bm{a}^0_{t}+\bm{b}_1,\\
& \bm{a}^2_{t} = \bm{W}_{2}\bm{a}^1_{t}+\bm{b}_2,\\
& \vdots \nonumber \\
& \bm{a}^{N_h}_{t} = \bm{W}_{N_h}\bm{a}^{N_h-1}_{t}+\bm{b}_{N_h},\\ 
& \bm{p}^{(m)} = \sigma_m(\bm{a}^{N_h}_{t}),
\label{eq_LSTM}
\end{eqnarray}
where $\bm{W}$ and $\bm{b}$ are weights and biases to be obtained through training. $\sigma$ is the sigmoid function. $N_b$ is the number of hidden layer. $\sigma_m$ is the softmax function. $\bm{p}^{(m)}$ records the estimated probability of each class based on inputs from the training sample $m$ and weights of the model, i.e., $\bm{p}^{(m)} = [p_{1}^{(m)}, p_{2}^{(m)},...,p_{N_c}^{(m)}]^T$. The loss function is represented by a cross entropy equation:
\begin{equation} 
\begin{split}
E^{m} = -\sum^{N_c}_{i=1} \gamma_{i} \log(p_{i}^{(m)}),
\end{split}
\end{equation}
where $\gamma_i$ is a binary indicator (0 or 1), with value $1$ if class label $i$ is the correct classification and $0$ otherwise. Thus, the training of the model is to minimize the sum of the entropy of the training set to find the best model parameters through backward propagation \citep{hecht1992theory}, i.e., 
\begin{equation} 
\min_{\bm{W},\bm{b}} \sum_m E^{m}. 
\end{equation}
Thus, adversary's behavior can be forecasted through
\begin{equation}
\bar{y}^r = \{ i \; | \; 1 \leq i \leq N_c, \forall 1 \leq j \leq N_c \; : p_i^{(m)} \geq p_j^{(m)}\},
\label{eq_forecast}
\end{equation}
where $\bar{y}^r$ is the predicted adversary's strategy, which can be used to calculate the possible adversary's dispatch order $\bar{\bm{v}}^r$ by using the upper bounds of the strategy shown in Tab.~\ref{tab_strategies}.  

\subsection{Dispatch Agent}

The goal of the convoy dispatch is to determine the desired attributes that need to be carried by a convoy to maximize the win rate. A convoy with higher attributes, especially in firepower, indicates a higher chance to win. However, as resources are limited, the less vehicles are ordered, the higher the chance that the corresponding order can be achieved by the base agent. Thus, the convoy dispatch should be carefully planned to guarantee the win rate of current mission without overusing resources.

To avoid the overuse or underuse of available attributes, it is important to have an accurate evaluation of the available vehicles before ordering vehicle convoy. In addition, time delays exist in operations, the dispatch order needs to be placed ahead of time to provide the base agent with enough time to prepare the order. Nevertheless, the actual planning is determined by base agent and the feasibility of the dispatch order can only be estimated based on the historical behaviors of base agent. This difficulty becomes one of the challenges in placing the dispatch order. Even we ascertain that our convoy order can be achieved by base agent, whether the dispatched convoy can win the event or not is still a question. As damage mechanism is sealed to both fleets, both of them need to speculate the damage mechanism based on the experience and estimate the probability of success accordingly. 

To resolve this problem, we decouple the estimation of event success in two parts, which are the feasibility of order $p^b_f$ and conditional success rate if order is feasible $p^b_s$. The probability of wining an event for a blue convoy $p^b_{w}$ can be calculated as
\begin{equation}
\begin{split}
p(\text{win}) & = p(\text{win}|\text{order is feasible}) p(\text{order is feasible})\\
p^b_{w} & = p^b_s p^b_f.
\end{split}
\label{eq_pspf}
\end{equation}

\subsubsection{Feasibility}

\begin{figure}[h!]
	\begin{center}
		\setlength{\unitlength}{0.012500in}%
		\includegraphics[scale=0.35]{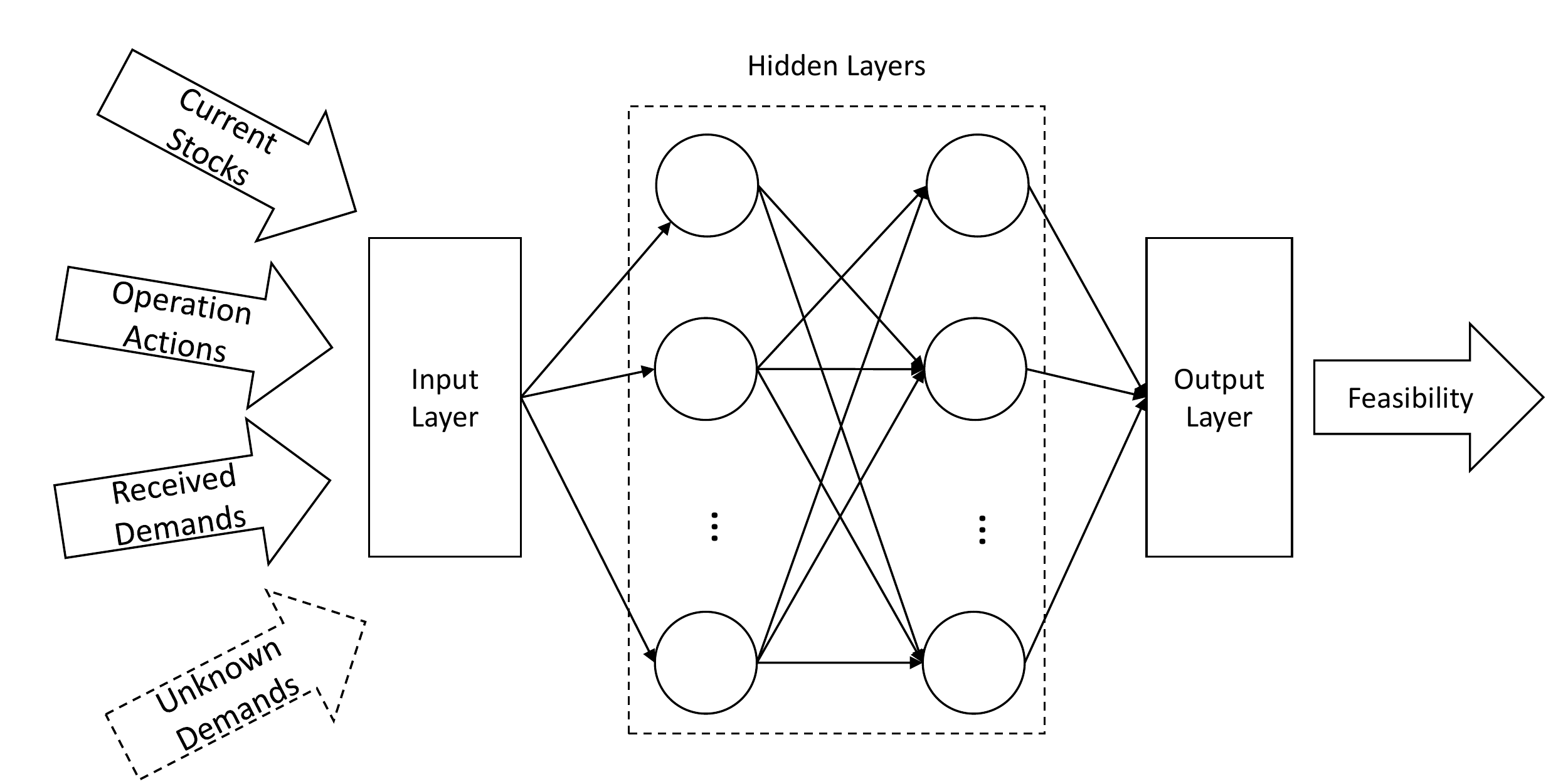}
	\end{center}
	\caption{Neural network model for feasibility}
	\label{fig_feasNN}
\end{figure}

The feasibility of the order can be defined as $0$ when infeasible and $1$ when feasible based on a comparison between the dispatch order $\bm{do}^b(t)$ and the actually dispatched convoy $\bm{o}^b_v (t)$. Specifically, an order is feasible if $\bm{do}^b(t) \leq \mathbf{M}_\mathrm{va} \bm{o}^b_v(t)$, and infeasible otherwise, where $\mathbf{M}_\mathrm{va}$ is the mapping between vehicles and attributes and inequality of two vectors refers to the entry-by-entry inequality of the values in the two vectors. The relationship between factors and feasibility is complex and nonlinear because optimizations are implemented in operation planning. We implement a neural network model \citep{hagan1996neural, atsalakis2018forecasting, rezaee2018integrating} to capture these nonlinear inter-connections, as shown in Fig.~\ref{fig_feasNN}. The output of the training set is the feasibility of the order (1 for feasible, 0 for infeasible).

With enough training, the model is capable of evaluating the feasibility of dispatch orders across diverse operation situations. To capture changes in the inventory operation strategy, we periodically re-train the model based on the latest operation information. The relationship between model inputs and feasibility rate can be described by
\begin{equation}
\begin{aligned}
p_f = f_f(\bm{do}^b(t), \bm{s}^b(t), \delta\bm{s}^b(t+1),..., \delta\bm{s}^b(t+T_p), \bm{D}(t)),
\end{aligned}
\label{eq_f}
\end{equation}
where, $\bm{s}^b(t)$ records the number of vehicles and component stocks on base at time $t$. $\delta\bm{s}(t+\tau)$ is the changes in inventory stocks at time $t+\tau$ from unfinished operational actions scheduled previously. 

\subsubsection{Conditional Success Rate}
%

\begin{figure}[h!]
	\begin{center}
		\setlength{\unitlength}{0.012500in}%
		\includegraphics[scale=0.35]{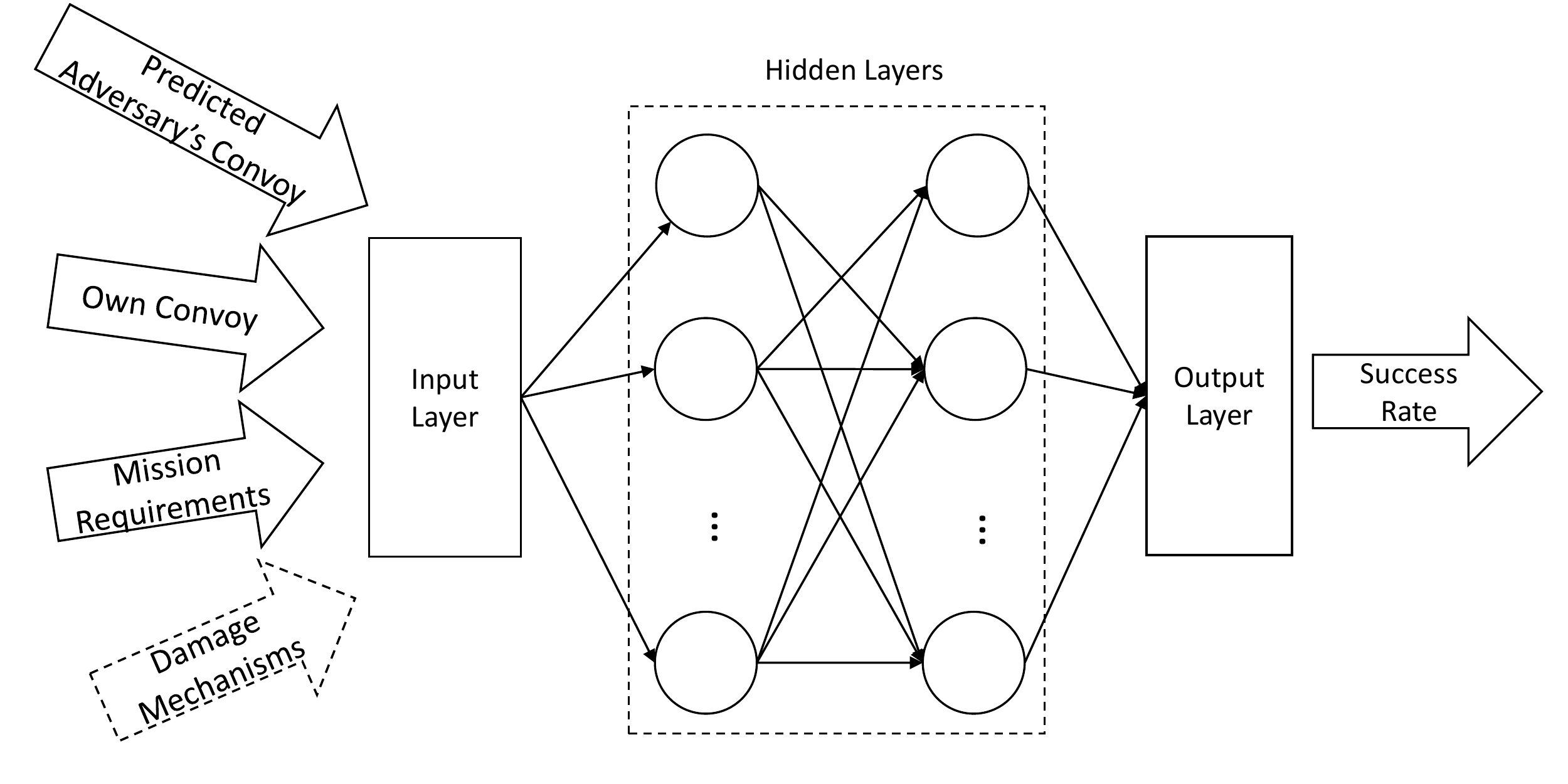}
	\end{center}
	\caption{Neural network model for success rate}
	\label{fig_succNN}
\end{figure}

Vehicle damage plays an important role in determining the success of a mission. However, vehicle damage is driven by a stochastic process and varies according to changes in terrain, commander's goal and military tactics. A model is needed to capture the complexity of the damage mechanism.  We adopted another neural network model for forecasting the success rate, as shown in Fig.~\ref{fig_succNN}.

The output of the training set are success history of previous events (1 for success, 0 for failure). Given forecasted adversary's convoy attributes $\bm{\bar{v}}^r(t)$, the trained model will yields the conditional win rate for a certain dispatch order and mission requirements. The model is capable of capturing changes in the damage mechanisms by continuously feeding in the latest event information and results. By denoting the trained neural network model for success as $f^b_s$, the probability of success can be calculated as 
\begin{equation}
p^b_s = f^b_s(\bm{do}^b(t), \bar{\bm{v}}^r(t), \bm{D}(t)).
\label{eq_s}
\end{equation}

\subsubsection{Optimization}

For each dispatch order $\bm{do}^b(t)$, the approach above provides a way to estimate the probability of success and probability of feasibility based on predicted adversary's behavior, demand information and inventory status. An optimization model can be used to seek the optimal dispatch order to maximize the win rate or minimize the failure rate, i.e., $J = 1- p_b^w$. Combining with Eqns.~\ref{eq_f} and \ref{eq_s}, a nonlinear programming model can be formulated to seek the optimal dispatch order as 
%
\begin{equation}
\begin{aligned}
& \min_{\bm{do}^b(t)} 
& & 1-f_f(\bm{do}^b(t), \bm{s}^b(t), \delta\bm{s}^b(t+1),..., \delta\bm{s}^b(t+T_p), \bm{D}(t)) f^b_s(\bm{do}^b(t), \bar{\bm{v}}^r(t), \bm{D}(t))  \\
& \text{s.t.} & &  (a) \; \bm{do}^b(t) \geq 0,\\
\label{eq_opt1}
\end{aligned}
\end{equation}
where $\bm{do}^b(t)$ is the decision variable that specifies the desired attributes to be carried by the blue convoy. Thus, the number of decision variables is the number of attribute types. However, it is intractable to obtain the global optimum by a gradient-based approach because of the non-convexity in objective function. Hence, in this study we implement a pattern search technique to yield optimal dispatch decisions. 

As the minimized failure rate can be any value in the range of $[0,1]$, the dispatch agent should be capable of giving up the mission once a very high failure rate is predicted. There is also a stream of literature studying risk preferences in repeated and evolutionary games \citep{roos2010risk, lam2006formalizing, zhang2018role}.  We define the $\epsilon_f$ as a customizable parameter to represent the minimal failure rate that can be tolerated, which modifies the dispatch order $\bm{do}^b(t)$ as 
\begin{equation}
\bm{do}^b(t)= \left\{
\begin{aligned}
&= \; \bm{0}, \;\;\; &\text{for}  \;\;\; (1-p_w)>\epsilon_f,\\
&= \; \bm{do}^b(t) &\text{for}  \;\;\; (1-p_w)\leq\epsilon_f.
\end{aligned}
\right.
\end{equation}

Thus, a convoy can be dispatched only when the probability of success is high enough. Risk aversion behavior is related to $\epsilon$ which is constant during operation. As a future work, it is also interesting to vary $\epsilon_f$ over time to seek an advanced fleet operation strategy, i.e., combination of risk-prone and risk-averse strategies\citep{roos2010risk}.

\subsection{Base Agent}

The base agent is the one to plan operational actions to accomplish the orders based on the behavior analysis from the inference agent and the dispatch order suggestion from the dispatch agent. Li and Epureanu proposed a model predictive control based approach to real-timely schedule the operation actions in reacting to the received demands\citep{li2018IABS}. However, they did not consider the possible damage that occurs during fleet operation. In this section, we further that research by considering the possible damage during fleet operation, and manage the inventory based on the resulting diverse conditions. For convenience, we simplify the notation of operation actions for fleet $x$ from $\bm{o}^x$ to $\bm{o}$ in this section as no adversary is considered.

It is important to schedule the operation actions properly to recover damaged resources and increase utility rate because resources for each player are limited and repairable. It is also essential to allocate the working capacity properly to balance between order satisfaction and damage recovery. In this section, we first model military fleet operation as a time-varying dynamical system. Then, a model predictive control is proposed to manage the system dynamics thus achieving overall operation management. 

\subsubsection{Dynamical System}
The dynamics of the fleet operation is mainly due to the changes of inventory stocks and remaining demands, in terms of
\begin{enumerate}
	\item {Vehicle stocks, $\bm{I}_v = [I_{v_1},I_{v_2},...,I_{N_v}] $}, $N_v$ being the total number of vehicle types,
	\item {Module/component stocks, $\bm{I}_c = [I_{c_1},I_{c_2},...,I_{N_c}] $}, $N_c$ being the total number of component types,
	\item {Damaged vehicles, $\bm{I}_{dv} = [I_{dv_1},I_{dv_2},...,I_{N_{dv}(t)}] $}, $N_{dv}(t)$ being the total number of damaged vehicles at time $t$,
	\item {Damaged components, $\bm{I}_{dc} = [I_{dc_1},I_{dc_2},...,I_{N_{c}}] $},
	\item {Unsatisfied demands, $\bm{I}_{a} = [I_{a_1},I_{a_2},...,I_{N_{a}}] $}.
\end{enumerate}

Although healthy vehicles and damaged vehicles are recorded in a similar vector, their meanings are different. For healthy stocks and damaged components, the subscript of variable is the type of vehicle/component; the value of variable indicates the number. For damaged stocks, the subscript is the index of the damaged vehicle, which is created based on the vehicle receipt date. Binary values are used to represent the status of a vehicle, where 1 represents that the damaged vehicle remains to be repaired; 0 indicates that the damaged stock is recovered or not received yet. For each damaged vehicles, information are recorded as 
\begin{enumerate}
\item{The vehicle type, $\bm{v}_{lt}=[v_{l1t},v_{l2t},...,v_{lN_vt}]$, where $v_{lkt} = 1$ if damaged vehicle of index $l$ is a vehicle in type $k$, and 0 otherwise,}
\item {The damaged components, $\bm{v}_{ldc}(t) = [{v}_{ldc_1}(t), {v}_{ldc_2}(t), ..., {v}_{ldc_{N_c}}(t)]$ in a damaged vehicle of index $l$, where ${v}_{ldc_i}$ is the number of damaged components of type $i$ in the damaged vehicle of index $l$,}
\item{The healthy components, $\bm{v}_{lc}(t)= [{v}_{lc_1}(t), {v}_{lc_2}(t), ..., {v}_{lc_{N_c}}(t)]$ in a damaged vehicle of index $l$, where ${v}_{lc_i}$ is the number of healthy components of type $i$ in the damaged vehicle of index $l$,.} 
\end{enumerate}

These data are time variant because the number and type of damaged vehicles keep changing with newly occurred damages and vehicle recoveries. We create a state for each newly arrived damaged vehicle and remove the corresponding state once the damaged vehicle is recovered, i.e., its state value changes from 1 to 0. To maintain a desirable resource utility rate, vehicle repair is usually proceeded in a short time after damaged is received. The real-time adjustment of $\bm{I}_{dv}$ keeps the size of vector short and bypass the numerous states incurred from diverse vehicle damage patterns, as mentioned in Section~\ref{problemStatement}. 

Vehicle conditions are reported to the base agent as one of the inputs. All inputs, $\bm{a}=[\bm{a}_r,\bm{a}_d,\bm{do}]^T$, to the fleet operation system are

\begin{enumerate}
	\item {Returning healthy vehicles, $\bm{a}_r = [a_{r_1},a_{r_2},...,a_{r_{N_v}}] $, where $a_{r_k}$ is the number of returned vehicles of type $k$,}
	\item {Returning damaged vehicles, $\bm{a}_d = [a_{d_1},a_{d_2},...,a_{d_{N_{dv}}}] $, where $a_{d_l}$ represents a newly arrived damaged vehicles of index $l$,}
	\item {Dispatch order from dispatch agent, $\bm{do} = [do_1,do_2,...,do_{N_a}]$, where $do_{k}$ is the number of vehicles of type $k$ ordered by the dispatch agent.}
\end{enumerate}

Based on the characteristics of fleet operation, the operational actions to be determined are also distinct. For conventional fleet, the operation actions include
\begin{enumerate}
	\item {Convoy dispatch, $\bm{o}_v = [o_{v_1},o_{v_2},...,o_{{v_{N_v}}}]$, where $o_{v_k}$ is the number of vehicles of type $k$ in the dispatched convoy,}
	\item {Recovery of damaged vehicle, $\bm{o}_{dr} = [o_{dr_1},o_{dr_2},...,o_{dr_{N_{dv}}}]$, where $o_{dr_l}$ indicates damaged vehicle of index $l$ to be repaired (1) or not (0),}
	\item {Recovery of damaged component, $\bm{o}_{c} = [o_{c_1},o_{c_2},...,o_{N_{c}}] $, where $o_{c_i}$ indicates the number of damaged modules of type $i$ to be repaired.}
\end{enumerate}

Given the time to accomplish the operational action $o_x$ as $\tau_x$, the dynamics of the vehicle stocks $I_{v_k}$ of type $k$, component stocks $I_{c_i}$ of type $i$, damaged vehicles $I_{dv_l}$ of index $l$, damaged components $I_{dc_i}$ of type $i$, and remaining attributes $I_{a_h}$ of type $h$ are governed by
\begin{equation}
I_{v_k}(t+1) = I_{v_k}(t) - o_{v_k}(t) + a_{r_k}(t) + \sum_{l=1}^{N_{dv}} v_{lkt}(t) o_{dr_l}(t - \tau_{v_l}),
\label{eq_conv1}
\end{equation}
\begin{equation}
I_{c_i}(t+1) = I_{c_i}(t) + o_{c_i}(t-\tau_{c_i}) - \sum_{l=1}^{N_{dv}} v_{lc_i}(t) o_{dr_l}(t),
\label{eq_conv2}
\end{equation}
\begin{equation}
 I_{dv_l}(t+1) = I_{dv_l}(t) - o_{dr_l}(t) + a_{d_l}(t),
\label{eq_conv3}
\end{equation}
\begin{equation}
I_{dc_i}(t+1) = I_{dc_i}(t) - o_{c_i}(t) + \sum_{l=1}^{N_{dv}} v_{ldc_i}(t) o_{dr_l}(t-\tau_{dr_l}),
\label{eq_conv4}
\end{equation}
\begin{equation}
 I_{a_h}(t) = do_{h}(t) - \sum_{k=1}^{N_v} M_{v_k a_h} o_{v_k}(t).
\label{eq_conv5}
\end{equation}
where $M_{v_ka_h}$ represents the amount of attributes of type $h$ carried by vehicle of type $k$. By introducing fleet modularity, several additional operation actions are available, in terms of
\begin{enumerate}
	\item {Vehicle assembly, $\bm{o}_a = [o_{a_1},o_{a_2},...,o_{a_{N_v}}] $, where $o_{a_k}$ indicates the number of vehicles of type $k$ to be assembled,}
	\item {Vehicle disassembly, $\bm{o}_d = [o_{d_1},o_{d_2},...,o_{d_{N_v}}] $, where $o_{d_k}$ indicates the number of vehicles of type $k$ to be disassembled,}
	\item {Vehicle reconfiguration, $\bm{o}_r = [o_{1 2},o_{1 3},...,o_{{N_v}{N_v - 1}}] $, where $o_{kk'}$ indicates the number of vehicles of type $k$ to be reconfigured to vehicles of type $k'$,}
	\item {Damaged vehicle disassembly, $\bm{o}_{dd} = [o_{dd_1},...,o_{dd_{N_{dv}}}]$, where $o_{dl}$ indicates the damaged vehicles of index $l$ to be disassembled.}
\end{enumerate}

With consideration of these actions, the governing equations of the dynamics become
\begin{equation}
\begin{split}
I_{v_k}(t+1)  = &  I_{v_k}(t)  - o_{v_k}(t) + a_{r_k}(t) + \sum_{{k'}\neq k}^{N_v}o_{k' k}(t-\tau_{k' k})  - \sum_{{k'}\neq k}^{N_v}o_{k k'}(t) \\ 
& + \sum_{l=1}^{N_{dv}} v_{lkt}(t)o_{dr_l}(t - \tau_{dr_l}) + o_{a_k}(t-\tau_{a_k}) -  o_{d_k}(t),
\end{split}
\label{eq_mod1} 
\end{equation}
\begin{equation}
\begin{split}
I_{c_i}(t+1) & =  I_{c_i}(t) + o_{c_i}(t-\tau_{c_i}) - \sum_{k=1}^{N_v} M_{v_k c_i} [o_{a_k}(t) -  o_{d_k}(t-\tau_{d_k}) \\
& + \sum_{{k'}\neq k}^{N_v}o_{k' k}(t) - \sum_{{k'}\neq k}^{N_v}o_{k k'}(t-\tau_{k k'})] + \sum_{l=1}^{N_{dv}} v_{lc_i}(t) o_{dd_l}(t - \tau_{dd_l}),
\end{split}
\label{eq_mod2}
\end{equation}
\begin{equation}
I_{dv_l}(t+1) = I_{dv_l}(t) - o_{dr_l}(t)  - o_{dd_l}(t) + a_{d_l}(t),
\label{eq_mod3}
\end{equation}
\begin{equation}
\begin{split}
I_{dc_i}(t+1) =  I_{dc_i}(t) - o_{c_i}(t) + &\sum_{l=1}^{N_{dv}} v_{ldc_i}(t) o_{dr_l}(t - \tau_{dr_l}) + \sum_{l=1}^{N_{dv}} v_{ldc_i}(t) o_{dd_l}(t - \tau_{dd_l}),
\end{split}
\label{eq_mod4}
\end{equation}
\begin{equation}
I_{a_h}(t) = do_{h}(t) - \sum_{k=1}^{N_v} M_{v_k a_h} o_{v_k}(t).
\label{eq_mod5}
\end{equation}
where $M_{v_kc_i}$ represents the number of components of type $i$ in vehicle of type $k$. Because of the delays in operation actions, current inventory stocks might be influenced by previously-determined actions. In other words, the current actions may impact the stock level in the future. Thus, we define the state of the system by all inventory statuses, $\bm{I}_{t}(t)=[\bm{I}_v, \bm{I}_c, \bm{I}_{dv}, \bm{I}_{dc}, \bm{I}_a]^T$, that might be influenced by current actions, $\bm{s}(t)$, i.e.,
\begin{equation}
\bm{s}(t) = [\bm{I}_{t}(t), \bm{I}_{t+1}(t), \bm{I}_{t+2}(t), \bm{I}_{t+3}(t)...\bm{I}_{t+\tau_{max}}(t)]^T.
\label{state}
\end{equation}

We use input matrices $\bm{B}_\tau(t)$ to connect the current actions at time $t$ to inventory level at a later time $t+\tau$. Furthermore, the damage in stocks keeps changing over time. The matrices that connect to previous states $\bm{A}(t)$, actions $\bm{B}_\tau(t)$ and inputs $\bm{C}(t)$ are also time-varying matrices. Thus, the system dynamics for both fleets can be written as
\begin{equation}
 \bm{I}(t+1) =  \bm{I}(t) + \sum_{\tau=0}^{\tau_{max}} \big[ \bm{B}^c_\tau (t) \bm{o}^c (t - \tau)\big] + \bm{C}^c (t)  \mathbf{a} (t- \tau) ,\label{eq_conv} 
\end{equation}
\begin{equation}
\bm{I}(t+1) =  \bm{I}(t) + \sum_{\tau=0}^{\tau_{max}} \big[ \bm{B}^m_\tau (t) \bm{o}^m (t - \tau)\big] + \bm{C}^m (t)  \mathbf{a} (t- \tau),
\label{eq_mod}
\end{equation}
where 
\begin{equation}
\bm{o}^c (t) = [\bm{o}_v(t), \bm{o}_{dr}(t),  \bm{o}_c(t)]^T,
\end{equation}
\begin{equation}
\bm{o}^m (t) = [\bm{o}_v(t), \bm{o}_{dr}(t), \bm{o}_c(t), \bm{o}_a(t), \bm{o}_d(t), \bm{o}_r(t), \bm{o}_{dd}(t)]^T.
\end{equation}
Thus, a state space model can be created to record the influence from the actions at a single time point to the states in the short future. For both fleets, the dynamical system can be represented by

\begin{equation}
\bm{s}(t+1) = \bm{A}(t)\bm{s}(t) + \bm{B}(t)\bm{o}(t) + \bm{C}(t)\mathbf{a}(t),
\label{eq_stockDyn}
\end{equation}
where \begin{equation}
\bm{A}(t) = \begin{bmatrix}
\bm{0}_{{n_s(t)}\times {n_s(t)}}      & \bm{I}_{{n_s(t)}\times {n_s(t)}} & \bm{0}_{{n_s(t)}\times {n_s(t)}} & \dots & \bm{0}_{{n_s(t)}\times {n_s(t)}} \\
\bm{0}_{{n_s(t)}\times {n_s(t)}}      & \bm{0}_{{n_s(t)}\times {n_s(t)}}  & \bm{I}_{{n_s(t)}\times {n_s(t)}} & \dots & \bm{0}_{{n_s(t)}\times {n_s(t)}} \\
\hdotsfor{5} \\
\bm{0}_{{n_s(t)}\times {n_s(t)}}      & \bm{0}_{{n_s(t)}\times {n_s(t)}} & \bm{0}_{{n_s(t)}\times {n_s(t)}} & \dots & \bm{I}_{{n_s(t)}\times {n_s(t)}} \\
\bm{0}_{{n_s(t)}\times {n_s(t)}}      & \bm{0}_{{n_s(t)}\times {n_s(t)}} & \bm{0}_{{n_s(t)}\times {n_s(t)}} & \dots & \bm{I}_{{n_s(t)}\times {n_s(t)}} \\
\end{bmatrix},
\end{equation}
\begin{equation}
\bm{B}(t) =  [\bm{B}_0(t),\bm{B}_1(t), \bm{B}_2(t), ... , \bm{B}_{\tau_{max}}(t)]^T, 
\end{equation}
\begin{equation}
\bm{C}(t) = [\bm{C}_0(t),\bm{C}_1(t), \bm{C}_2(t), ... , \bm{C}_{\tau_{max}}(t)]^T.
\end{equation}

\subsubsection{System Control}

The goal of system control is to meet the received dispatch orders on time. In the decision-making process, predictions of future system states are always involved. For example, given several dispatch orders, one may want to know what are the influences from satisfying one order on satisfying others. Compared to classical control methodologies, e.g., PID control, model predictive control (MPC) makes better use of future information and adapts to the system changes \citep{li2017robustness}. We separate this section into two parts where we discuss future state prediction first and optimization of operation decisions second.

\subsubsection*{Future State Prediction}

Because of time delays in the operation actions, the operation decisions made at the current time have to guarantee the match between the attributes of the dispatched convoy and the ordered attributes. Given
\begin{enumerate}
	\item Current system states $\bm{s}(t)= [\bm{s}(t+1), \bm{s}(t+2), \bm{s}(t+3), ... ,\bm{s}(t+t_p)]^T$ , where $t_p$ is the planning horizon,
	\item Operation actions in the future $\underrightarrow{\bm{o}}_{t} = [\bm{o}(t), \bm{o}(t+1), \bm{o}(t+2), ... ,\bm{o}(t+t_p-1)]^T$,
	\item System input, $\underrightarrow{\mathbf{a}}_{t} = [\mathbf{a}(t), \mathbf{a}(t+1), \mathbf{a}(t+2), ... ,\mathbf{a}(t+t_p-1)]^T$,
\end{enumerate}

The future system states $\underrightarrow{\bm{s}}_{t+1}$ are predictable by iteratively using Eq.~\ref{eq_stockDyn}. Thus, one can express $\underrightarrow{\bm{s}}_{t+1}$ as a function of $\underrightarrow{\bm{o}}_{t}$ as

\begin{equation}
\underrightarrow{\bm{s}}_{t+1} = \bm{P}(t) \bm{s}(t) + \bm{H}(t) \underrightarrow{\bm{o}}_{t} + \bm{G}(t) \underrightarrow{\mathbf{a}}_{t} \label{eq_future}
\end{equation}
\begin{equation}
 \bm{P}(t) =  [\bm{A}(t),\bm{A}^2(t), \bm{A}^3(t), ... , \bm{A}^n(t)], 
\end{equation}
where
\begin{equation}
\bm{H}(t)=
\begin{bmatrix} \bm{B}(t)  & \bm{0}& \dots &\bm{0}\\
\bm{A}(t)\bm{B}(t) & \bm{B}(t) & \dots &\bm{0}\\
\hdotsfor{4}\\
\bm{A}(t)^{t_p-1}\bm{B}(t)&\bm{A}^{t_p-2}(t)\bm{B}(t)& \dots &\bm{B}(t)\\
\end{bmatrix},
\end{equation}

\begin{equation}
\bm{G}(t)=
\begin{bmatrix} \bm{C}(t)  & \bm{0}& \dots &\bm{0}\\
\bm{A}(t)\bm{C}(t) & \bm{C}(t) & \dots &\bm{0}\\
\hdotsfor{4}\\
\bm{A}(t)^{t_p-1}\bm{C}(t)&\bm{A}^{t_p-2}(t)\bm{C}(t)& \dots &\bm{C}(t)\\
\end{bmatrix},
\end{equation}
with $\bm{P}(t)$ being the matrix that connects the future system outputs with current system states. $\bm{H}(t)$ and $\bm{G}(t)$ being the matrix that connects the system outputs with the future operation actions and inputs respectively. The dynamic model keeps updating to ensure we optimize the operation actions based on the most recent system status.

\subsubsection*{Cost Function}

The optimization of the fleet operation originates from two facts: 1. a convoy with insufficient attributes suffers a remarkable risk of mission failure; 2. a convoy with redundant attributes can also deteriorate the overall fleet performance from utility reduction. Furthermore, we consider several operational costs that may be significant in realistic fleet operations. As a summary, the costs of interest are

\begin{enumerate}
	\item {Attribute redundancy cost for type $h$ attribute, $c_{o_h}$,}
	\item {Attribute insufficiency cost for type $h$ attribute, $c_{u_h}$,}
	\item {ADR action cost: assembly of vehicle of type $k$, $c_{a_k}$, disassembly of vehicle of type $k$, $c_{d_k}$, vehicle reconfiguration from type $k$ to type $k'$, $c_{kk'}$,}
	\item {Recovery cost: repair a component of type $i$ $c_{c_i}$, repair a damaged vehicle of index $l$, $c_{dr_l}$, disassembly a damaged vehicle of index $l$, $c_{dd_l}$,}
	\item {Inventory holding cost for holding a healthy/damaged component $c_{hc_i}/c_{hv_k}$, and a healthy/damaged vehicle $c_{hdc_i}, c_{hdv_l}$.}
\end{enumerate}

The cost function can be expressed as
\begin{equation}
\begin{aligned}
J & & =  \sum_{\tau=t}^{\tau=t+T_p-1} & \Big[\sum_{h=1}^{N_a} c_{o_h} I_{a_h}^+(\tau) + \sum_{h=1}^{N_a} c_{u_h} I_{a_h}^-(\tau) + \sum_{l=1}^{N_{dv}} (c_{dr_l} o_{dr_l}(\tau) + c_{dd_l} o_{dd_l}(\tau)) \\
& & & + \sum_{k=1}^{N_v} [c_{a_k} o_{a_k}(\tau) + c_{d_k} o_{d_k}(\tau)) + \sum_{k'\neq k}^{N_v} c_{k k'} o_{k k'}(\tau)] + \sum_{i=1}^{N_c} c_{c_i} o_{c_i}(\tau)\Big]\\
& & + \sum_{\tau=t+1}^{\tau=t+T_p} & [ \sum_{i=1}^{N_c} c_{hc_i} I_{c_i}(\tau) + \sum_{k=1}^{N_v} c_{hv_k} I_{v_k}(\tau)+ \sum_{i=1}^{N_c} c_{hdc_i} I_{dc_i}(\tau) + \sum_{l=1}^{N_{dv}}c_{hdv_l} o_{dv_l}(\tau)], \\
\end{aligned}
\end{equation}
where $I_{a_h}^+(\tau)$ and $I_{a_h}^-(\tau)$ are non-negative auxiliary variables representing the positive part and negative parts of $I_{a_h}(\tau)$, which satisfies
\begin{equation}
I_{a_h}(\tau) = I_{a_h}^+(\tau) - I_{a_h}^-(\tau).
\end{equation}

We record the insufficient and the redundant attributes during the planning horizon as $\underrightarrow{\bm{I}^+_{a}}$ and $\underrightarrow{\bm{I}^-_{a}}$ respectively. The holding costs and the actions-related costs are also aggregated as $\underrightarrow{\bm{c}_h}$ and $\underrightarrow{\bm{c}_a}$. By substituting in Eq.~\ref{eq_future}, we create a mixed-integer programming model to optimize operational decisions as
\begin{equation}
\begin{aligned}
& \min_{\underrightarrow{\bm{I}^+_{a}},\underrightarrow{\bm{I}^-_{a}}, \underrightarrow{\bm{o}}_{t}}
& & \bm{c}_{o} \underrightarrow{\bm{I}_{a}}^+ + \bm{c}_{u} \underrightarrow{\bm{I}_{a}}^- + \underrightarrow{\bm{c}_h} \bm{X}_s[\bm{P}(t)\bm{s}(t)  + \bm{H}(t) \underrightarrow{\bm{o}}_{t} ] \\
& \text{s.t.} & & (a) \; \underrightarrow{\bm{o}}_{t} \geq 0 \;\; \text{and integer} \\
& & & (b) \; \bm{X}_s \underrightarrow{\bm{s}}_{t+1} \geq 0  \\ 
& & & (c) \; \sum \bm{o}(t) \leq \bar{P} , \quad \forall t  \\
& & & (d) \; \underrightarrow{\bm{I}^+_{a}} - \underrightarrow{\bm{I}^-_{a}} = \bm{X}_a[\bm{P}(t) \bm{s}(t) + \bm{H}(t) \underrightarrow{\bm{o}}_{t} + \bm{G}(t)\underrightarrow{\mathbf{a}}_{t}]\\
& & & (e) \; o_{dd_l}(t) + o_{dr_l}(t) \leq 1 , \quad \forall t,\\
\end{aligned}
\end{equation}
where $\bm{X}_s$ and $\bm{X}_a$ record indices of inventory stocks and remaining dispatch orders in the created state structure respectively. Constraint (a) ensures that all operational decisions are non-negative and integer; (b) indicates that the amount of inventory stocks are non-negative; (c) ensures that the ADR actions are always constrained by the maximum action capacity $\bar{P}$; (d) preserves the balance between auxiliary variables and remaining orders to be satisfied; (e) specifies that each damaged vehicle can only be recovered by one recovery strategy. As the cost function and constraints are linear and the number of decision variables is large, we first implement a cutting-plan to reduce the decision space and then use an integer programming solver to obtain the solution. The time required for decision-making at each time point is less than 1 second for operating 5 types of modular vehicles with a planning horizon of 12 hours.

\section{Numerical Illustrations}
In this section, we provide numerical illustrations for a generalized mission scenario to study the different impacts of modularity on fleet performance. In general, it may be difficult to know all fleet parameters accurately. However, it may be possible to obtain reasonable estimates for these parameters. In this study, we assume the resources provided to the fleet operation are fixed and equal, which can be imagined as a competition of two fleets on an isolated island. One of them is a conventional fleet; the other is a modular fleet. Initially, ten of each type of vehicles and components are provided to both fleets. Demands randomly occur at the battlefield based on a Poisson distribution with a time interval of 10 hours. Demands include personnel capacity $d_p$, material capacity $d_m$ and firepower $d_f$, which are generated based on a Gaussian distribution as shown in Eq.~\ref{eq_demand1}, \ref{eq_demand2}, \ref{eq_demand3}. 
\begin{eqnarray}
& d_p \sim \mathcal{N}(40,\,15) 
\label{eq_demand1}\\
& d_m \sim \mathcal{N}(50,\,20) 
\label{eq_demand2}\\
& d_f \sim \mathcal{N}(30,\,10) 
\label{eq_demand3}
\end{eqnarray}

Because of the lack of diversity in the existing designs of modular vehicles, we borrow five types of modular vehicles as well as six types of modules from \citep{li2017robustness}. The attributes carried by each vehicle are summarized in Tab.~\ref{tab_v2a}.

\begin{table}[h!]
	\caption{Mapping between vehicles and attributes}
	\label{tab_v2a}
	\begin{center}
		\begin{tabular}{c c c c c c}
			& & \\ 
			\hline
			\hline
			Vehicle Type& $1$ & $2$ & $3$  & $4$ & $5$   \\
			\hline
			\hline
			Firepower & $1$ & $3$ & $8$  & $0$ & $6$ \\
			Material capacity & $2$ & $6$ & $2$  & $2$ & $8$ \\
			Personnel capacity & $4$ & $1$ & $0$  & $10$ & $5$ \\
			\hline
		\end{tabular}
	\end{center}
\end{table}

The costs of insufficiency and redundancy are created based on heuristic estimates. For example, convoys usually suffer a high risk of failure when the attributes of the dispatched convoys are less than the attributes ordered. Thus, the cost for attribute insufficiency is much higher than the cost of attribute redundancy, i.e., $c_{o_h} \gg c_{u_h}, \; \forall h=1,2,...,N_a$. The costs for operation actions are created based on their difficulty and the time required for their execution. 

We assign the time required for module assembly and disassembly as constant vectors $\mathbf{\tau}_\mathrm{ma}$ and $\mathbf{\tau}_\mathrm{md}$. Vehicle assembly and disassembly times $\mathbf{\tau}_\mathrm{va}$ and $\mathbf{\tau}_\mathrm{vd}$ are calculated by summing up all the times required for assembly of all its components. Similarly, for repair and reconfiguration, we sum all the time needed for actions to process each individual component in the vehicle. We assume that the interfaces between components are well-designed to achieve quick vehicle reconfiguration, where assembly and disassembly time for all types of components are of 1 hour and of 0.5 hour respectively. We assume that ADR actions are executed at working stations. Thus, the number of stations determines the amount of available working capacity. In this study, the number of available work stations for both fleets is 15.

We execute a discrete event model to simulate the fleet competition for three years. We separate the 3-year mission into two parts, namely the stochastic stage ($1^{st}$ year), and the learning stage ($2^{nd}$ and $3^{rd}$ year). In the stochastic stage, the dispatch agent randomly chooses a dispatch strategy based on Tab.~\ref{tab_strategies} and passes this decisions to the base agent. First-year operations generate time-series data, including event history and feasibility records, which are important inputs for the learning models. Training of the learning models for each fleet starts at the beginning of the learning stage, where the inference agent and the dispatch agent begin to make decisions based on the historical adversary's behavior. Learning models are also updated monthly to ensure they reflect the adversary's latest behaviors.

\section{Fleet comparison}

\begin{figure} [h!]
	\begin{center}
		\setlength{\unitlength}{0.012500in}%
		\includegraphics[scale=0.55]{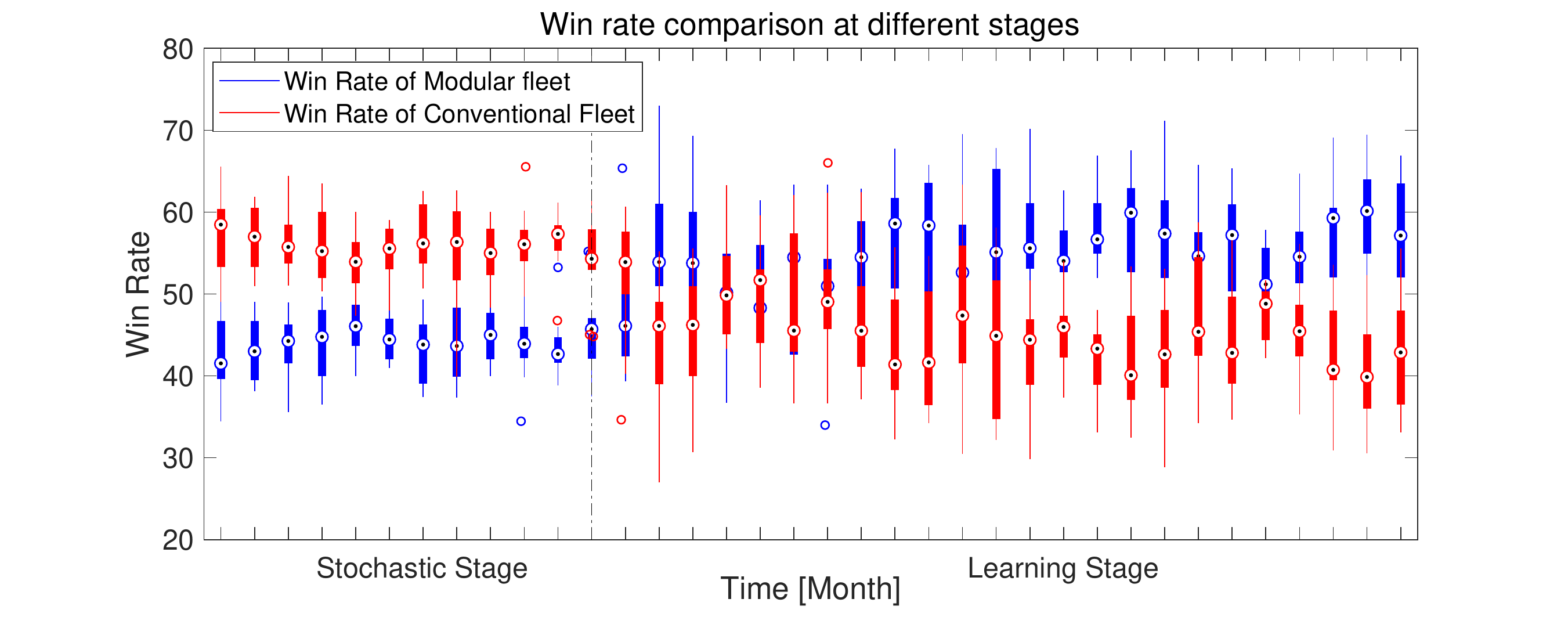}
	\end{center}
	\caption{Comparison of win rate between modular fleet and conventional fleet}
	\label{fig_winComp}
\end{figure}

In this section, we compare the performance of the modular fleet and the conventional fleet in the two stages. As one of the important metrics in measuring the fleet performance, we first compare the probability of winning based on the results from multiple simulations. These results are shown in Fig.~\ref{fig_winComp}. Conventional fleet outperforms the modular fleet during the stochastic stage. However, once the intelligence of the agents is introduced, i.e., once both fleets enter the learning stage, the modular fleet wins more often. To explain these results, we first compare the order success, and then compare the feasibility the attributes carried by the actual dispatched convoys of each of the fleets, damage suffered, and the estimation accuracy in inference of the adversary.

\begin{figure}[h!]
	\begin{center}
		\setlength{\unitlength}{0.012500in}%
		\includegraphics[scale=0.60]{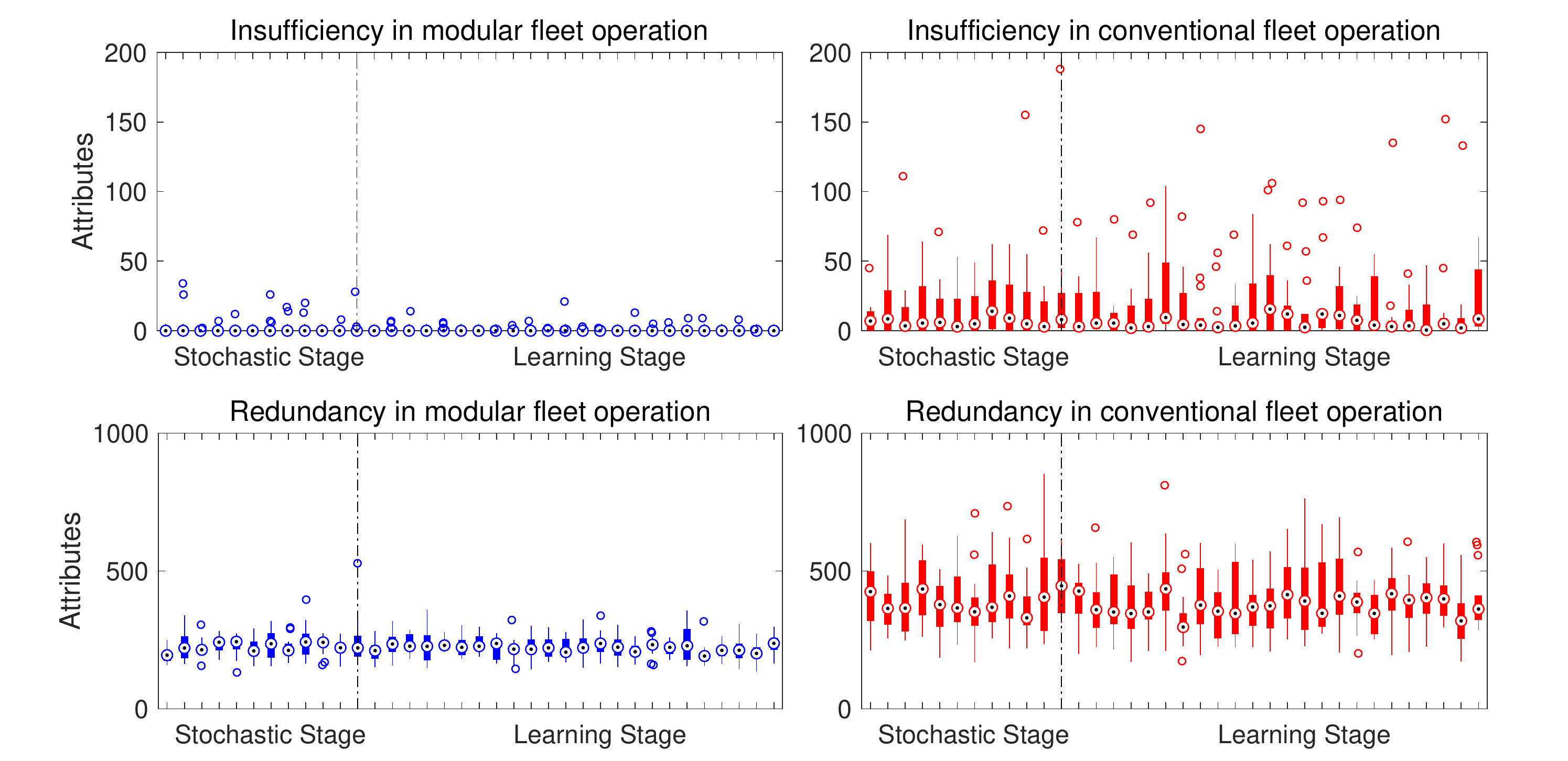}
	\end{center}
	\caption{Comparison of total mismatched convoy attributes in all types}
	\label{fig_errorComp}
\end{figure}

During the stochastic stage, dispatch agents from both fleets place dispatch orders based on a randomly selected strategy. The strategy selection and the order achievement are same for both fleets. To explain the better performance of the conventional fleet, it is necessary to explore the accuracy of both fleets in satisfying the dispatch orders. Mismatched attributes dramatically change the fleet performance: convoys with insufficient attributes may significantly raise the failure rate and the damage; convoys with redundant attributes may increase the win rate slightly, but they can also contribute to insufficiency in the short future because of limited resources. Thus, we calculate the amount of overused and insufficient convoy attributes during every month and compare the dispatch accuracy  in Fig.~\ref{fig_errorComp}.

Compared to the modular fleet, the conventional fleet suffers remarkable redundancy. The higher redundancy comes from the rigidity of the conventional fleet operation. The modular fleet can real-timely reconfigure itself to fit the dispatch orders. However, the conventional fleet can only wait for vehicles to return from the battlefield or for damaged vehicles to be recovered. This limitation hampers the ability of the conventional fleet to satisfy the dispatch orders. Once proper vehicles are scarce, the conventional fleet has to use improper vehicles with little desired attributes to avoid insufficiency. This rigidity in fleet operation is beneficial in improving the success rate during the stochastic stage, because the adversary cannot be aware of the unexpected additional attributes. However, once the adversary starts to learn the behavior of the conventional fleet, this advantage no longer exists. From failures during the stochastic stage, the modular fleet learns that the conventional fleet intents to dispatch convoys with superfluous attributes. As a solution, modular fleet increases the attributes of their dispatch orders correspondingly. These redundant attributes in conventional convoys are powerless in reacting to the intelligent response of the modular fleet and the conventional fleet cannot stay ahead.

\begin{figure}[h!]
	\begin{center}
		\setlength{\unitlength}{0.012500in}%
		\includegraphics[scale=0.55]{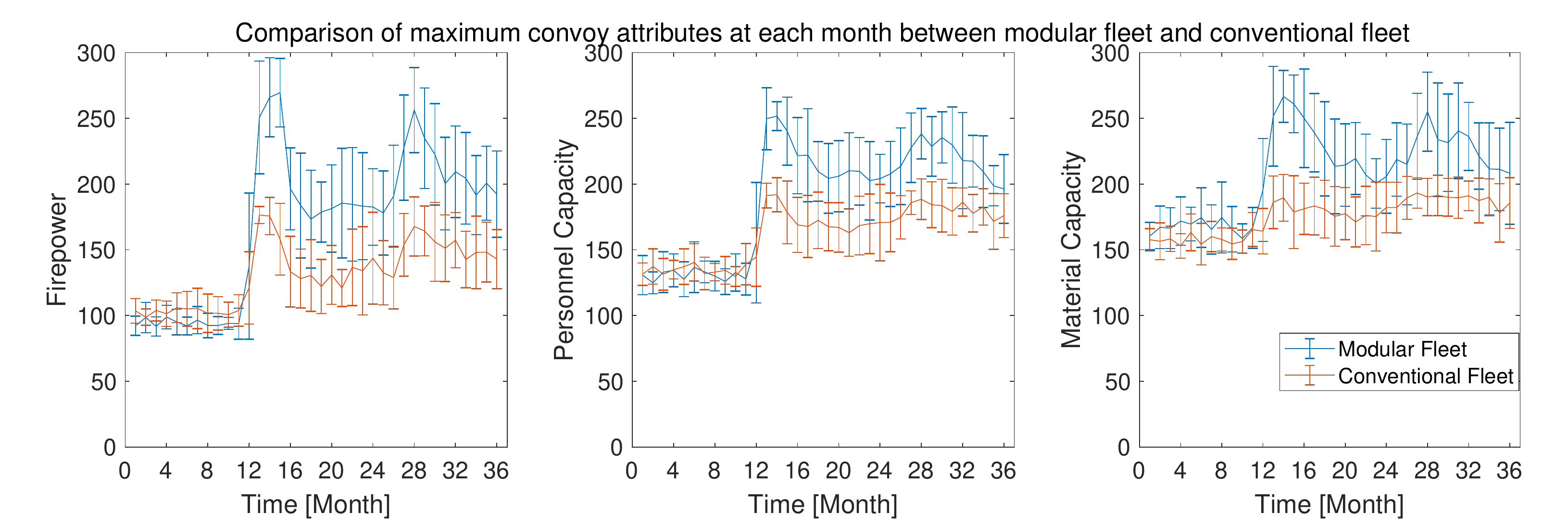}
	\end{center}
	\caption{Maximum value of attributes carried by convoys in different months}
	\label{fig_attriComp}
\end{figure}

Besides a better understanding of adversary's behavior, intelligent agents also improve their understanding of the game over time. To explore this process, we first compare the maximum values of convoy attributes dispatched in each month, and compare them of both fleets in Fig.~\ref{fig_attriComp}. After entering the learning stage, both fleets raise their dispatched attributes in all types, especially in firepower. Both fleets learn that the sufficient delivered attributes, especially for firepower, are the key to win the event through the trained success rate model $f_s^b$. With additional flexibility in fleet operation, modular fleet can always form a convoy that could provide required attributes with high safety level, which may not be achievable by the conventional fleet. 

\begin{figure}[h!]
	\begin{center}
		\setlength{\unitlength}{0.012500in}%
		\includegraphics[scale=0.60]{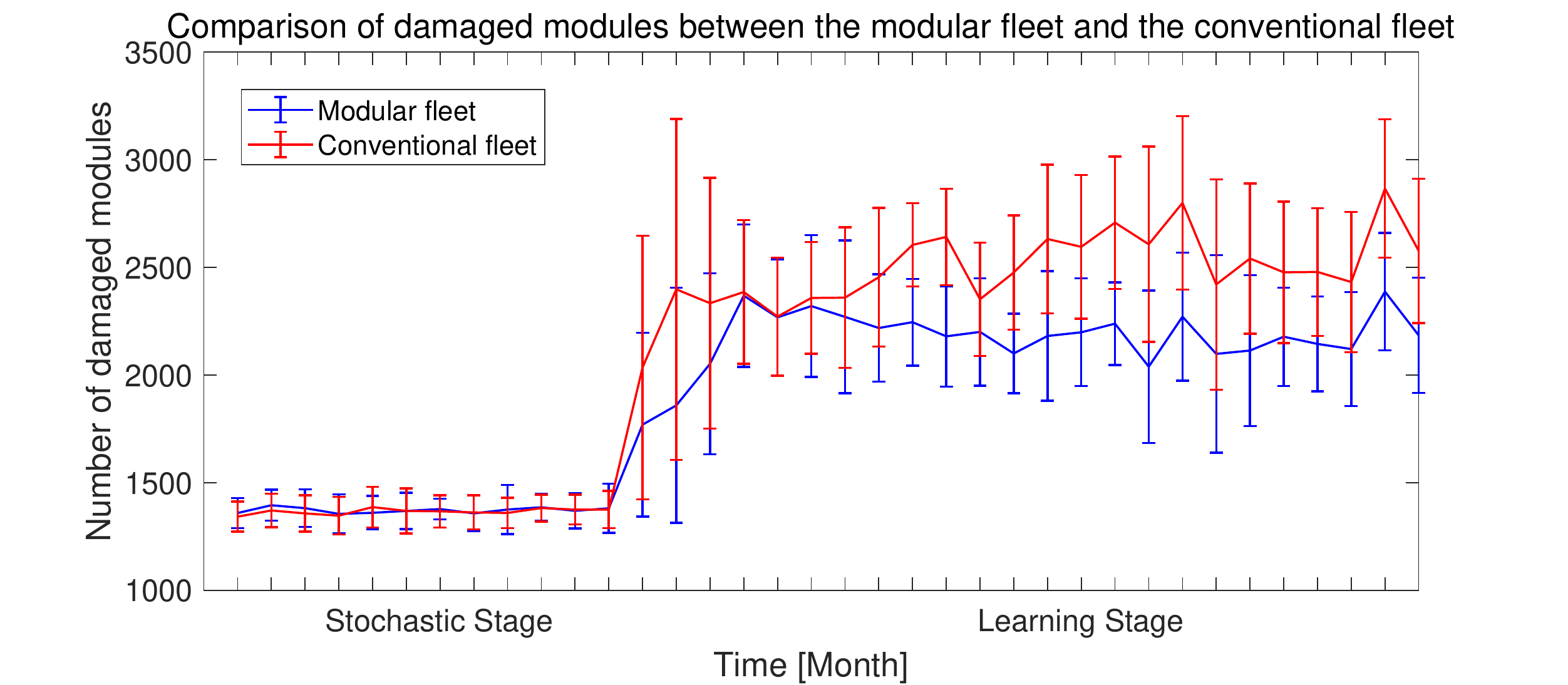}
	\end{center}
	\caption{Module damages occur during each month}
	\label{fig_dmgComp}
\end{figure}

The swift reconfiguration of the modular fleet leads to a dramatic increase in the damage to the adversary in the first few months of the learning stage, as shown in Fig.~\ref{fig_dmgComp}. Although the conventional fleet increases the firepower to fight back, the limitation in the vehicle structure results in a lower upper limit in the firepower. Thus, the difference in the fleet ability makes conventional fleet suffer higher damage from more dispatch orders, which forces the conventional fleet to operate in sub-healthy conditions for a long time.

\begin{figure}[h!]
	\begin{center}
		\setlength{\unitlength}{0.012500in}%
		\includegraphics[scale=0.6]{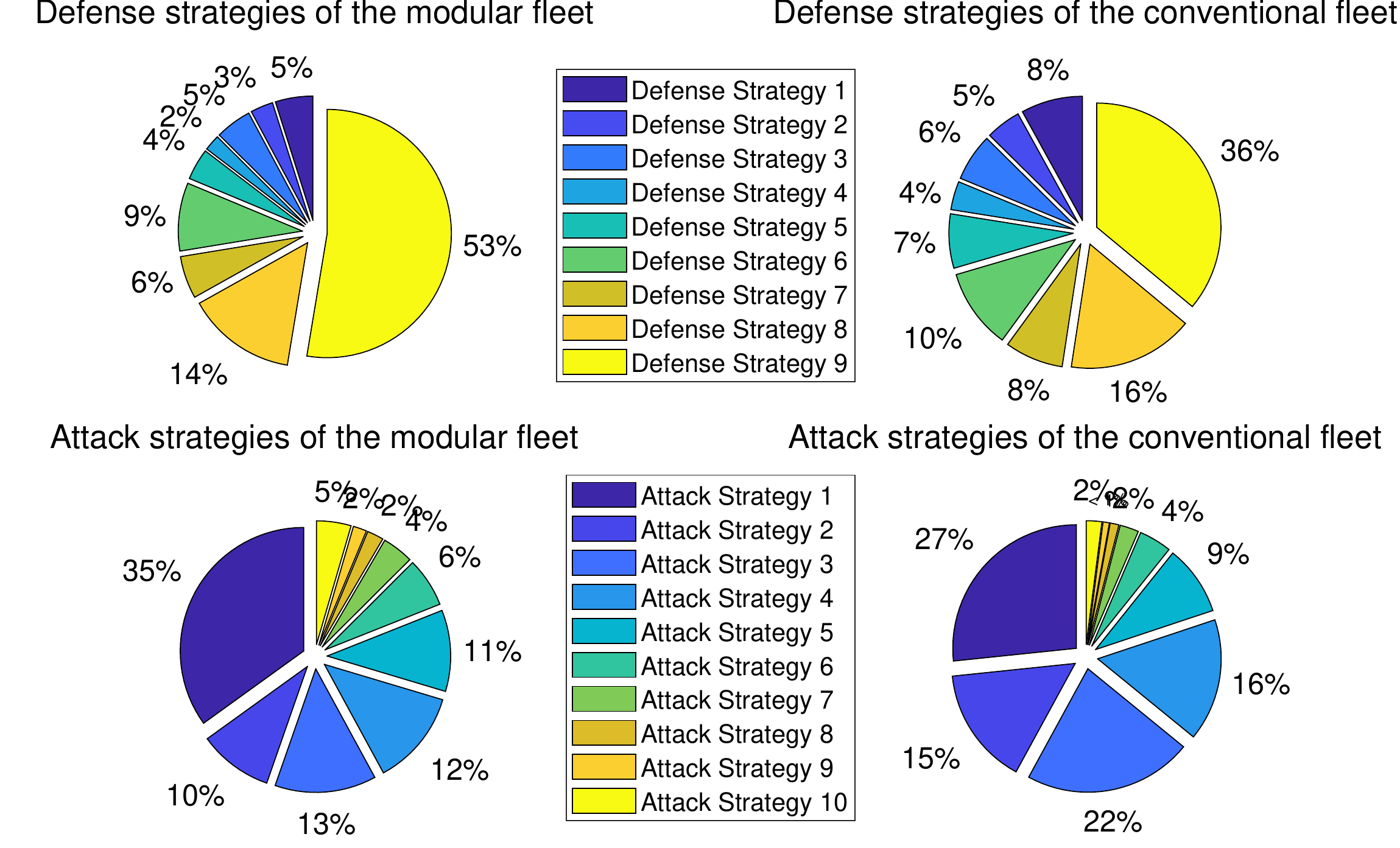}
	\end{center}
	\caption{Proportion of the attack and defense strategies during different stages}
	\label{fig_strategy}
\end{figure}

The strategies used in the learning stage are also distinct between the two fleets. Fig.~\ref{fig_strategy} compares the proportion of strategies adopted by the two fleets. After learning, both fleets prefer to select the defense strategy with large amount of firepower and a fair amount of capacity, i.e., strategies 8,9. Because of the flexibility of the fleet structure, the modular fleet can more easily adapt to the vehicle damage and the adversary's behavior, which leads to a better balance between different types of vehicles to perform a stronger strategy. The defense strategy selection also impacts the attack strategy. Compared to the modular fleet, the conventional fleet is much more likely to give up missions because of resource insufficiency. This weakness makes the modular fleet confident in dispatching little or even no combat vehicles and win events. As evidence, the proportion of strategy 1 used by the modular fleet is much higher than that by the conventional fleet. In addition, the modular fleet is more capable of performing aggressive strategies, i.e., strategies 8, 9 and 10, more often than the conventional fleet once a strong adversary is learned.  

\begin{figure}[h!]
	\begin{center}
		\setlength{\unitlength}{0.012500in}%
		\includegraphics[scale=0.60]{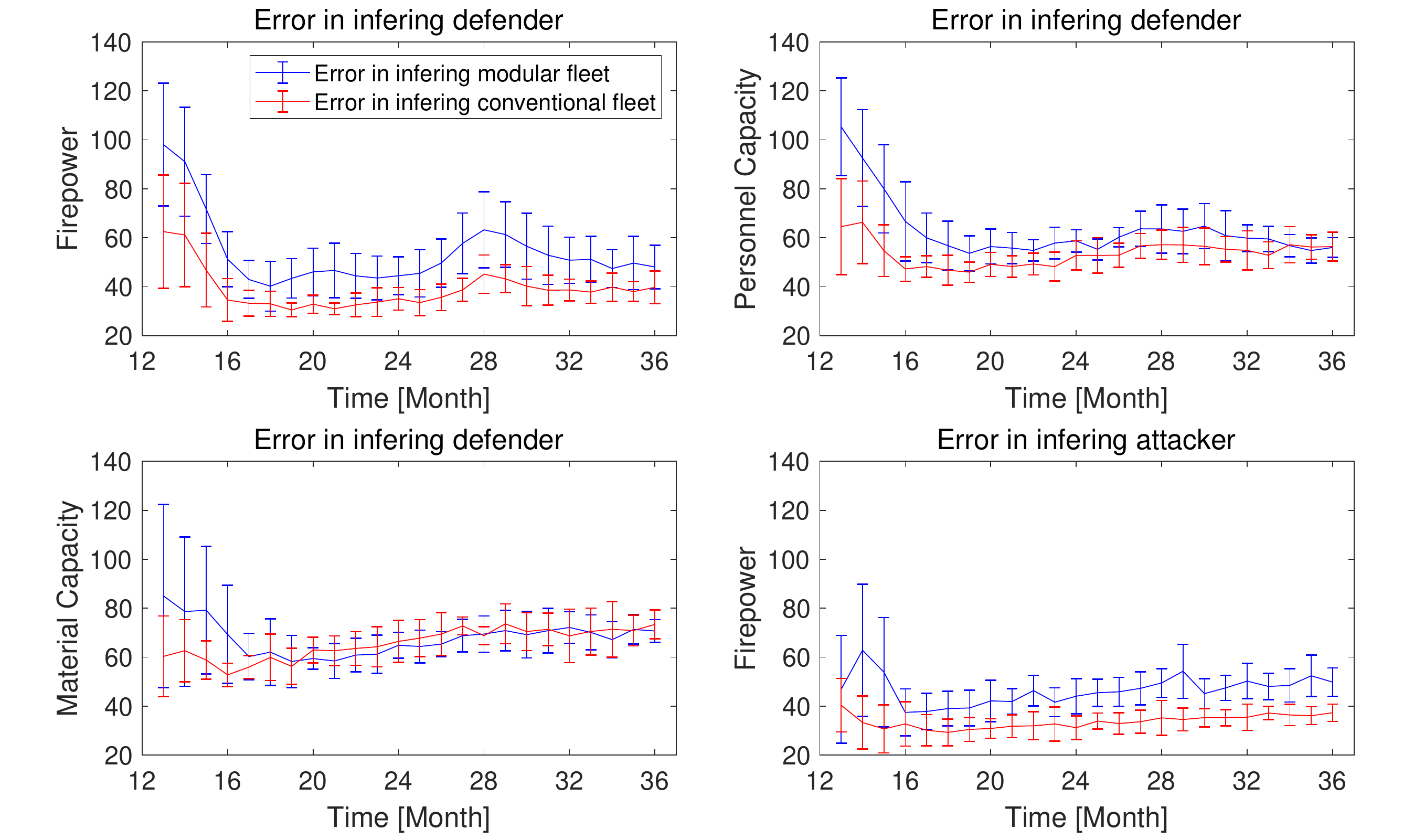}
	\end{center}
	\caption{Comparison of inference accuracy between the modular fleet and the conventional fleet}
	\label{fig_infComp}
\end{figure}

To further investigate the improved performance of the modularized fleet, we also compared the inference accuracy between the two fleets. We choose the mean square error (MSE) between forecasted and actual convoy attributes as the metric to quantify the inference accuracy. As can be seen from the comparison in Fig.~\ref{fig_infComp}, inference errors are significantly higher at the beginning of the learning stage, because agents are trained by the data from the stochastic dispatch, which contributes little to forecasting the behavior of an intelligent adversary. Once both fleets enter learning stage, the dispatch convoy orders made by trained learning models, $f_f$ and $f_s$, make adversary's behaviors are more explainable. As a result, inference errors are significantly reduced in the following four months. However, the inference errors keep fluctuating during the rest of the learning stage, because both fleets keep checking and countering each other's behavior. 

The results also show that it is easier to infer the strategy of the conventional fleet than that of the modular fleet, especially in the attribute of firepower. This phenomenon originates from the higher freedom in decision-making due to modularity. As a defender, a fleet usually needs to prepare a convoy with all types of attributes to satisfy the demands. With limited vehicle stocks, the decision maker of the conventional fleet has constrained choices of strategy. However, for the modular fleet, the decision maker can vary the dispatch strategy by real-time vehicle reconfiguration, i.e., reconfigure vehicles to switch from strategy 4 (higher safety stocks) to strategy 8 (more firepower). 

\begin{figure}[h!]
	\begin{center}
		\setlength{\unitlength}{0.012500in}%
		\includegraphics[scale=0.60]{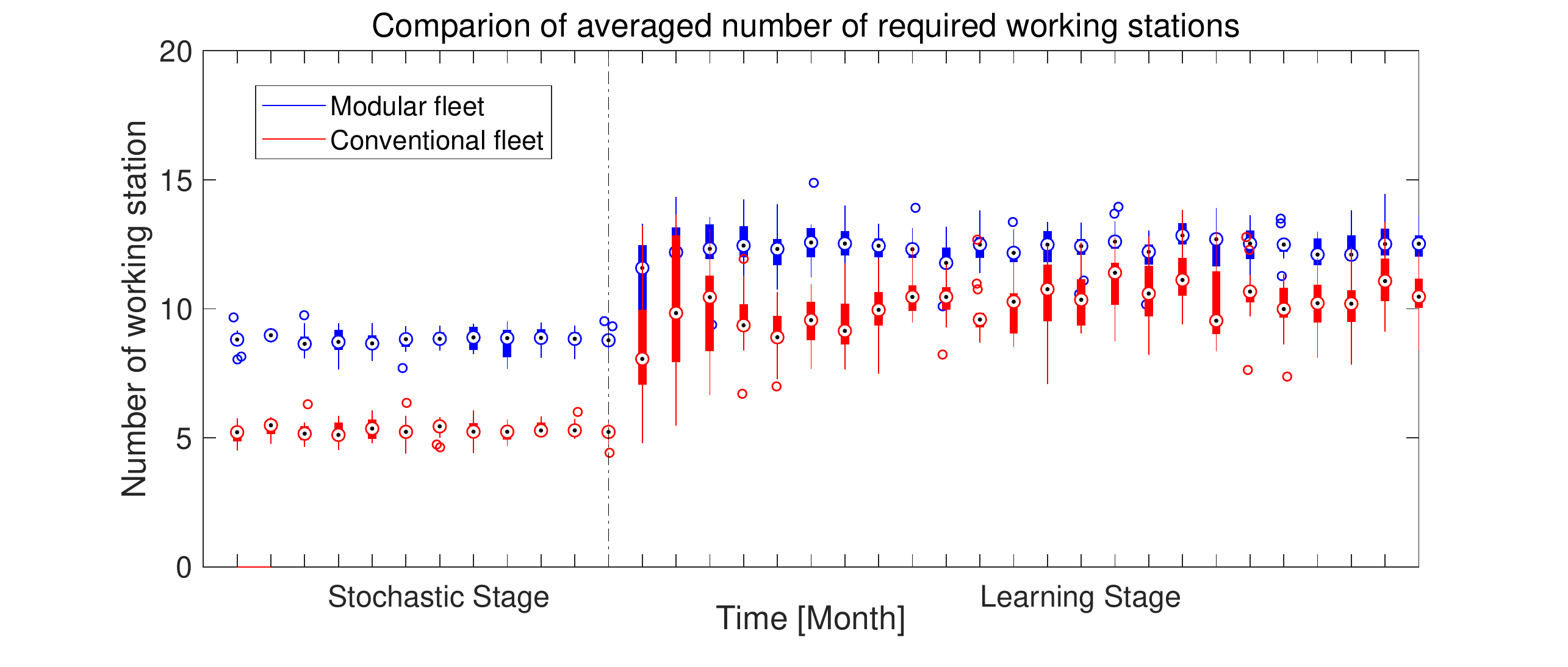}
	\end{center}
	\caption{Comparison of machine requirements between modular fleet and conventional fleet}
	\label{fig_capaComp}
\end{figure}

However, the burden of modularity is also significant, which is the high acquisition of capacity. According to Fig.~\ref{fig_capaComp}, the modular fleet always requires more working resources, i.e., personnel and assembly machine, than conventional fleet because of additional ADR actions. It can also be observed that the required working resources increase significantly once entering the learning stage, which are due to damages from smarter strikes by the adversary. The higher losses in the conventional fleet also shrink the difference in resource requirement at learning stage. In this study, we only tested the fleet performance at a certain capacity. Studies investigating on the influence of capacity can be found in the literature \citep{li2017robustness, li2018IABS}.

\section{Conclusions}
In this paper, we investigated the benefits and burdens from fleet modularization by simulating an attacker-defender game between a modular fleet and a conventional fleet. A noval intelligent agent based approach to proposed for battle-field decision-making process by combining the optimization and machine learning techniques. With continuous retraining, the model is capable to capture the evolution in strategies as a reaction to the adversarial behavior, which reveals game-theoretical behaviors.

We simulated the fleet competition for three years which are divided into a stochastic stage and a learning stage. By contrasting the simulation results from the two fleets, we found that the conventional fleet leads when both fleets are selecting strategies stochastically; the modular fleet outperforms the conventional fleet once the intelligence of the decision makers is considered. With additional operational flexibility from assembly, disassembly and reconfiguration actions, the modular fleet exhibits a better adaptability to adversarial actions, a stronger convoy, and a more significant unpredictability from the additional flexibility in operation.

\section*{Acknowledgment}
	This research has been supported by the Automotive Research Center, a US Army Center of Excellence in Modeling and Simulation of Ground Vehicle Systems, headquartered at the University of Michigan. This support is gratefully acknowledged. The authors are solely responsible for opinions contained herein. 

\bibliography{mybibfile}
\bibliographystyle{apa}
\end{document}